\documentclass[preprint,12pt,authoryear]{elsarticle}
\usepackage[margin=1in]{geometry}
\usepackage{amsmath,amssymb}
\usepackage{url,times,array,booktabs,tabularx,multirow,graphicx,float,placeins,indentfirst}
\usepackage[table]{xcolor}
\definecolor{TableBest}{HTML}{D1E7F5}
\definecolor{TableSecond}{HTML}{D6EDD8}
\usepackage[ruled,linesnumbered]{algorithm2e}
\SetKwComment{Comment}{// }{}
\SetAlCapFnt{\footnotesize}
\SetAlCapNameFnt{\footnotesize}
\usepackage{xspace}
\AddToHook{env/figure/begin}{\setlength{\abovecaptionskip}{6pt}}
\AddToHook{env/figure*/begin}{\setlength{\abovecaptionskip}{6pt}}
\newcommand{\method}{CoopDistill\xspace}
\DeclareRobustCommand{\rev}[1]{#1}
\DeclareRobustCommand{\revexp}[1]{#1}
\makeatletter
\AtBeginDocument{\ifx\curr@math@size\@empty
  \global\let\curr@math@size\f@size\fi}
\makeatother
\journal{Transportation Research Part C: Emerging Technologies}

\begin{document}
\raggedbottom
\begin{frontmatter}
\title{\method: \rev{Equilibrium-Guided} Multi-Agent Policy Distillation for Cooperative Driving}
\author[tongji]{Shiyu Fang}
\author[sjtu]{\rev{Sheng Zhao}}
\author[tongji]{Jiarui Zhang}
\author[tongji]{Chengkai Xu}
\author[tongji]{Xiaocong Zhao}
\author[tongji]{Peng Hang}
\author[tongji]{Jian~Sun\corref{corresponding}}
\cortext[corresponding]{Corresponding author. Email: sunjian@tongji.edu.cn}
\address[tongji]{Key Laboratory of Road and Traffic Engineering,
Ministry of Education, College of Transportation Engineering, Tongji University,
Shanghai 201804, China}
\address[sjtu]{\rev{School of Mechanical Engineering, Shanghai Jiao Tong University,
800 Dongchuan Road, Shanghai 200240, China}}

\begin{abstract}
\rev{Cooperative driving extends autonomous driving from individual vehicle
decisions to joint maneuvers that improve traffic safety and efficiency.
Optimization-based methods impose safety constraints but often fix
paths to simplify multi-vehicle planning, limiting spatial flexibility.
Multi-agent reinforcement learning (MARL) explores alternatives through
interaction, but coordinated learning and safe execution remain
challenging. Their integration nevertheless faces
challenges in exploration efficiency, experience allocation, and
execution reliability. To address these challenges, this paper proposes
CoopDistill, an equilibrium-guided multi-agent policy distillation
framework. A potential-game (PG) formulation provides a fixed-path
longitudinal reference, around which a residual MARL policy learns
longitudinal acceleration, conflict-approach timing, and lateral path
corrections. A hard-case curriculum concentrates experience on
interactions that PG leaves unresolved or substantially delayed.
A comparative rollout selector evaluates the joint consequences of
learned corrections against PG before choosing the executed action.
Experiments in merging, roundabout, and intersection scenarios compare
CoopDistill with six baselines at four connected automated vehicle
penetration levels. Relative to PG, success improves by 2.1--27.1
percentage points and delay decreases by 7.3--27.3\%, with no observed
collisions. Code is available at
\url{https://github.com/FanGShiYuu/CoopDistill}.}
\end{abstract}

\end{frontmatter}

\section{Introduction}

\rev{As autonomous driving moves toward commercial deployment, vehicles must negotiate shared road space at intersections, roundabouts, and merging areas \citep{schwarting2018planning,guanetti2018control}. In these complex conflict areas, decisions made from an individual vehicle's perspective may resolve an immediate encounter without coordinating the progression of the interacting group \citep{rios2017survey}. Vehicle connectivity creates opportunities for connected autonomous vehicles (CAVs) to exchange motion states and maneuver intentions and coordinate when and where to proceed \citep{talebpour2016influence,wang2020organization}. Cooperative driving can therefore complement individual autonomy by organizing joint maneuvers aimed at safer and more efficient traffic operation \citep{rios2017survey,malikopoulos2018decentralized}.}

\rev{Cooperative-driving methods coordinate vehicle decisions through optimization, multi-agent reinforcement learning (MARL), or combinations of the two. Optimization and game-theoretic methods represent interactions through explicit objectives and constraints, but joint motion and path planning becomes increasingly demanding as interactions multiply \citep{malikopoulos2018decentralized,katriniok2022distributed,wang2021game}. MARL provides maneuver flexibility through environmental interaction, although coordinated exploration and reliable execution remain challenging \citep{lowe2017multiagent,kiran2022deep,gronauer2022survey}. Hybrid methods seek to combine structured coordination with learned flexibility \citep{hewing2020learning,zanon2021safe,bautista2022autonomous,chen2026game,chen2026personalized,zeng2026reinforced}. The central question is how to learn useful improvements beyond the optimization reference while retaining its coordination benefits during execution.}

\begin{figure*}[t]
\centering
\includegraphics[width=0.9\textwidth]{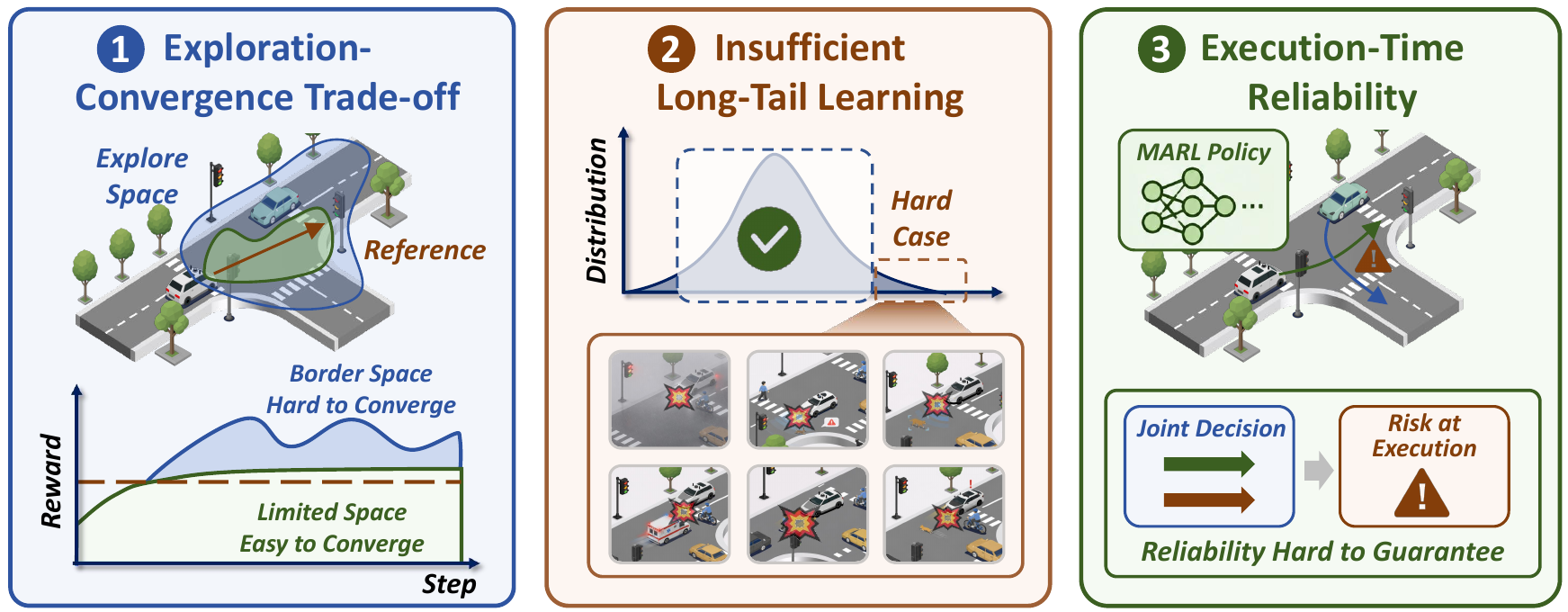}
\caption{\rev{Three challenges in integrating optimization and MARL.}}
\label{fig:CoopDistill_challenge}
\end{figure*}

\rev{Effective integration therefore involves three connected challenges (Fig.~\ref{fig:CoopDistill_challenge}). \textit{1) Exploration--Convergence Trade-off:} Exploring complete longitudinal and lateral actions across several vehicles creates a large joint search space and makes coordinated policies difficult to learn. Confining exploration to the optimizer's fixed-path decisions reduces that space but also excludes the path changes needed to release spatial obstruction. The challenge is to use the optimization solution to guide exploration without inheriting its maneuver restrictions \citep{johannink2019residual,gronauer2022survey}.}
\rev{\textit{2) Insufficient Long-Tail Learning:} Fixed-path coordination already resolves many routine encounters, leaving limited improvement to be learned from repeatedly visiting them. By contrast, infrequent interactions involving blocked paths or chains of yielding can leave an entire group of vehicles waiting. These cases require coordinated changes in both motion and path geometry, yet can occupy only a small share of training experience. Identifying such interactions and allocating sufficient experience to them is essential for learning capabilities beyond those of the reference \citep{bengio2009curriculum,narvekar2020curriculum}.}
\rev{\textit{3) Execution-Time Reliability:} Reward-based training does not explicitly enforce the coupled safety constraints of every new interaction \citep{wang2026feasible}. A learned correction may advance one vehicle while closing another's available gap, or remain collision-free while reducing collective progress. Execution therefore requires a state-dependent choice between the learned joint proposal and the optimization reference, based on their predicted feasibility and traffic consequences \citep{garcia2015safe,brunke2022safe}.}

\begingroup
\clubpenalty=10000
\rev{To address these challenges, we propose \method, an equilibrium-guided multi-agent policy distillation framework for cooperative driving. A potential-game (PG) formulation first produces a longitudinal coordination reference by maximizing a common potential over fixed-path acceleration sequences. Conditioned on this reference, a shared residual actor generates longitudinal acceleration, conflict-approach timing, and lateral path corrections, extending coordination from fixed-path motion to joint longitudinal--lateral maneuvers. Rollout-evaluated corrections provide action-level guidance, and multi-agent proximal policy optimization (MAPPO) refines the policy through vehicle interactions. A hard-case curriculum emphasizes scenes that PG cannot clear or clears only after substantial delay. At execution, a comparative rollout selector evaluates the learned joint proposal and PG from the same state, accepting the proposal when it satisfies the feasibility and utility conditions and otherwise retaining PG. The framework thus connects equilibrium-guided exploration, difficulty-directed training, and reference-based execution.}
\par
\endgroup

The contributions of this work are summarized as follows:
\begin{itemize}
\item \rev{An equilibrium-guided residual MARL formulation is developed with three action channels for longitudinal acceleration, conflict-approach timing, and lateral path adjustment. The shared policy learns joint corrections around a PG reference and extends the available cooperative maneuvers.}
\item \rev{A PG-based hard-case curriculum is constructed by ranking training scenes according to collision, completion, and delay outcomes. Sampling from the difficult subset together with fresh scenes directs experience toward interactions poorly resolved by the reference.}
\item \rev{A comparative rollout selector is formulated to evaluate learned and PG joint decisions from the same traffic state. Feasibility checks and a utility comparison determine which decision is executed, and the same selection process is included during training.}
\item \rev{A three-scenario evaluation compares the framework with six baselines across four CAV penetration levels. Component ablations and trajectory analyses examine its performance and the mechanisms by which coordinated path corrections release blocked traffic.}
\end{itemize}

\section{Related Work}

\rev{Cooperative driving methods differ in how joint vehicle decisions are represented and obtained. This section reviews optimization-based coordination and MARL, followed by hybrid methods that combine explicit planning with learned decisions.}

\subsection{Optimization-Based Methods}

\rev{Optimization-based cooperative driving determines the decisions of interacting vehicles by solving objectives subject to motion and interaction constraints. Reservation and auction mechanisms coordinate access to shared space by allocating occupancy intervals or passing priorities \citep{dresner2008multiagent,carlino2013auction,mirheli2018signal,li2019temporal}. Optimal-control and model predictive control (MPC) formulations additionally determine acceleration or trajectory profiles under safety, comfort, and efficiency requirements \citep{ntousakis2016optimal,malikopoulos2018decentralized,zhang2019decentralized,katriniok2022distributed,karimi2020cooperative}. Distributed methods exchange planned trajectories \citep{mirheli2019consensus}, while signal--trajectory and corridor optimization extend coordination across several locations \citep{feng2018spatiotemporal,guo2019joint,yu2019corridor,wang2020organization}. Recent lane-free formulations also optimize spatial use beyond conventional lane-following decisions \citep{malcolm2025lanefree,chi2024spatiotemporal}. These developments broaden the decisions coordinated, while increasing the number of coupled variables and constraints.}

\rev{Game-theoretic coordination makes the relationship between individual preferences and collective decisions explicit. A Nash equilibrium characterizes a joint decision from which no modeled participant benefits by changing its own action alone; this provides an interpretable account of yielding and proceeding in mixed traffic. Bargaining, coalition, and equilibrium formulations have been used to allocate priorities and coordinate maneuvers \citep{wang2021game,cao2025conflict,chen2026game}. Potential games align unilateral utility changes with a common potential \citep{monderer1996potential,fabiani2020multi}; \citet{fang2026adaptive} develop an adaptive PG approach for mixed-traffic cooperation and verify it through field tests. Equilibrium-based interpretation depends on the represented preferences and available actions. In particular, fixed-path formulations reduce computational complexity by restricting geometric choices, but this restriction can leave beneficial spatial rearrangements outside the feasible maneuver space.}

\subsection{MARL Methods}

\rev{MARL treats cooperating vehicles as agents whose policies are optimized through repeated interaction with the traffic environment. Agents observe local traffic, choose actions, and receive individual or shared rewards that reflect the resulting vehicle interactions. Value-based methods learn action values over discrete choices \citep{rashid2018qmix}, whereas actor--critic methods learn parameterized policies and value estimates for joint outcomes \citep{lowe2017multiagent,foerster2018counterfactual}. Under centralized training with decentralized execution (CTDE), joint observations support value estimation during training while each vehicle generates actions from its own policy input \citep{gronauer2022survey,yu2022surprising}. Shared policies and team rewards allow cooperation to be learned across a group of vehicles rather than prescribed through a fixed passing order.}

\rev{Driving applications use these approaches for unsignalized intersection coordination \citep{wang2023control} and joint longitudinal--lateral maneuvers in mandatory lane changing \citep{wang2024multiagent}. Simulation platforms support training over varied traffic conditions \citep{wu2022flow,vinitsky2018benchmarks}. Nevertheless, increasing the number of agents expands the joint action space and changes the environment experienced by each learner, complicating credit assignment and convergence. Policies trained for a particular geometry and interaction distribution must also be evaluated when those conditions change \citep{kiran2022deep,gronauer2022survey,wong2023challenges,gorsane2022standardised}. The challenge is therefore to obtain coordinated maneuver flexibility without requiring the policy to discover every useful traffic interaction from an unconstrained search.}

\subsection{Hybrid Methods}

\rev{Hybrid methods connect learning and optimization through different interfaces. Learning-based MPC adapts prediction or control parameters while retaining constrained optimization \citep{hewing2020learning,zanon2021safe}. In a parallel architecture, \citet{bautista2022autonomous} combine RL navigation with MPC and switch commands according to safety conditions at unsignalized intersections. Hierarchical learning-and-game formulations allocate cooperative partners and generate vehicle maneuvers at different decision levels \citep{chen2026game}. Policy distillation transfers decision examples into a policy \citep{rusu2016policy}, and residual RL represents learned corrections around an existing controller \citep{johannink2019residual}. Shielding and predictive filters further connect policy outputs to constrained execution \citep{alshiekh2018safe,fisac2019general,wabersich2021predictive}.}

\rev{These interfaces support adaptation and safety intervention, but the provision of an optimizer alone does not specify which joint corrections should be learned, which unresolved interactions deserve training, or whether a feasible correction improves collective passage relative to the available reference.}

\section{Problem Formulation}
\begingroup

\rev{Consider a road interaction area containing vehicles
\(\mathcal N=\{1,\ldots,N\}\), including CAVs
\(\mathcal C\subseteq\mathcal N\) and human-driven vehicles (HDVs).
Each vehicle has a prescribed origin--destination route whose local path
can be adjusted without changing its destination. Vehicle numbers, routes,
initial motion states, and CAV membership vary across scenes.
At decision instant \(t\), the joint state \(X_t\) contains vehicle motion,
route progress, and committed paths; \(\mathcal C_t\) indexes the CAVs
participating in the current interaction. Their joint decision
\(u_t=\{(a_{i,t},P_{i,t})\}_{i\in\mathcal C_t}\) specifies acceleration
requests and paths. HDVs follow their own driving behavior and are not
controlled by the cooperative policy. The system evolves as}
\begin{equation}
X_{t+1}=F(X_t,u_t;\xi_t),
\label{eq:transition}
\end{equation}
\rev{where \(F\) applies vehicle dynamics and physical execution constraints,
and \(\xi_t\) collects the HDV responses during the decision interval.}

Cooperation concerns both when vehicles occupy shared space and where
their paths overlap. Longitudinal decisions regulate approach times,
whereas local path adjustments can create space around an obstruction.
The objective is to promote collective progress and route completion,
reduce unresolved waiting, and avoid unsafe interactions. Let \(O_t\) denote the joint information
available to the cooperative decision system. The problem is formulated as
\begin{equation}
\begin{aligned}
\Pi^\star&\in\arg\max_\Pi
\mathbb E_\Pi\!\left[\sum_{t=0}^{T-1}\gamma^t R(X_t,u_t,X_{t+1})\right],\\
u_t&=\Pi(O_t),
\end{aligned}
\label{eq:task}
\end{equation}
where \(\Pi\) denotes the \rev{joint cooperative policy}, \(T\) is the
episode horizon in decision steps, and \(\gamma\in(0,1]\) is the
discount factor. The stage return \(R\) rewards traffic progress and
vehicle completion while penalizing unresolved traffic and unsafe outcomes.
Executed motion is subject to acceleration, speed, road-boundary, steering,
and lateral-acceleration limits, while collision avoidance additionally
depends on coupled vehicle interactions. Section~\ref{sec:methodology}
\rev{develops an equilibrium-guided framework that combines a PG coordination
reference with learned motion and path adjustments and compares their
predicted consequences before execution.}
\endgroup

\section{Methodology}
\label{sec:methodology}
\begingroup
\raggedbottom

\begin{samepage}
\rev{\method builds learned maneuvers around a PG coordination reference
(Fig.~\ref{fig:coopdistill_framework}). The reference supplies longitudinal
coordination, while a residual MARL policy learns joint motion and path
corrections to release spatial obstruction. To develop these corrections
where they are most needed, a hard-case curriculum mixes PG-identified
difficult scenes with fresh scenes during training. The resulting policy
proposes candidate maneuvers through the decoder; their effects are then
evaluated against PG through finite-horizon rollouts, so execution retains
the reference whenever a candidate fails the feasibility or utility
conditions.}
\par
\end{samepage}

\begin{figure*}[t]
\centering
\includegraphics[width=0.95\textwidth]{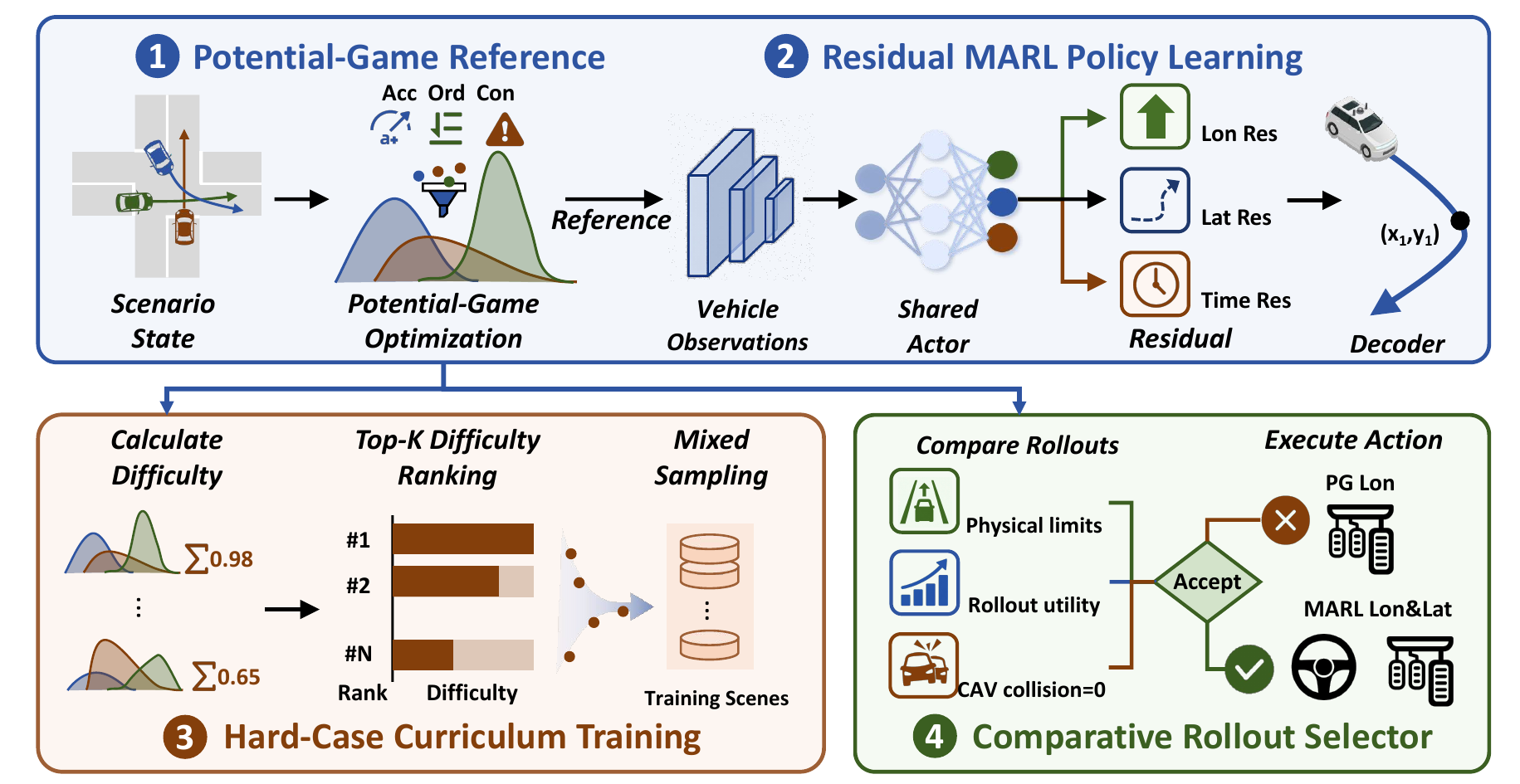}
\caption{\rev{Overview of \method. A PG coordination reference guides residual
MARL learning, hard cases direct training experience, and rollout comparison
selects the executed joint decision.}}
\label{fig:coopdistill_framework}
\end{figure*}

\subsection{Potential-Game Reference}
\label{sec:reference}

Longitudinal coordination must balance a vehicle's progress with the
space required by others approaching the same conflict region. Potential
games organize these interdependent preferences through a common
potential \citep{monderer1996potential,fabiani2020multi}. We construct
the PG reference from acceleration preferences, conflict proximity, and
estimated passing order, then optimize acceleration sequences subject
to motion and stopping constraints. Each solve holds the committed paths
fixed: these are the prescribed paths in PG-only operation and may include
earlier accepted corrections in \method.

For each \rev{participating CAV} \(i\in\mathcal C_t\), let
\(\boldsymbol a_i\in\mathbb R^{H_P}\) be its acceleration sequence over
an \(H_P\)-step prediction horizon, and let \(A\) collect these sequences.
The potential combines an acceleration incentive with directed conflict
interactions. Using discount factor \(\gamma_P\in(0,1]\) and vector
entries \(w_k=\gamma_P^k\), \(k=0,\ldots,H_P-1\), it is written as
\begin{equation}
\begin{aligned}
 \Phi_P(X_t,A)&=\sum_{i\in\mathcal C_t}\boldsymbol c_i^\top\boldsymbol a_i,\\
 \boldsymbol c_i&=\boldsymbol w+B_P^\top\boldsymbol w
             \sum_{j\in\mathcal I_i(t)}(q_{ji}-q_{ij}),\\
 q_{ij}&=\sigma_{ij}
 (d_{\rm near}-|d_{ij,i}-d_{ij,j}|)(d_{\rm near}-d_{ij,j}),
\end{aligned}
\label{eq:potential}
\end{equation}
where \(\mathcal I_i(t)\) contains vehicles whose current paths conflict
with that of \(i\), and \(d_{ij,i}\) and \(d_{ij,j}\) are their respective
distances to the corresponding geometric conflict stations. The matrix
\(B_P=\Delta_P^2L^2\), with prediction interval \(\Delta_P\) and
lower-triangular matrix of ones \(L\), maps acceleration sequences to the
travel approximation used in the potential. With interaction distance
\(d_{\rm near}>0\), a directed term \(q_{ij}\) is active only when both
vehicles \rev{remain in the interaction} and both distances lie in \((0,d_{\rm near})\);
otherwise it is zero. The sign
\(\sigma_{ij}\) equals 1 when \(d_{ij,i}/\max(v_i,v_\epsilon)\)
exceeds \(d_{ij,j}/\max(v_j,v_\epsilon)\), and equals \(-1\) otherwise,
where \(v_\epsilon>0\) regularizes low speeds.

The first term in \(\boldsymbol c_i\) encourages acceleration, while
the incoming and outgoing conflict terms adjust that incentive according
to conflict proximity and estimated passing order. Both directed terms
are retained for each pair. HDVs affect these interaction coefficients
through their predicted arrival times but are not decision variables.
Because the coefficients are evaluated at \(X_t\) and held fixed during
the solve, \eqref{eq:potential} is a linear, state-conditioned coordination
objective; the conflict geometry is updated at the next decision instant.

{
The potential gives the acceleration subproblem an equilibrium interpretation.
With the current state and interaction coefficients fixed, define the marginal
payoff \(U_i(X_t,A)=\boldsymbol c_i^\top\boldsymbol a_i\).
For any feasible unilateral change from \(\boldsymbol a_i\) to
\(\boldsymbol a_i'\), with the other sequences \(A_{-i}\) held fixed,
\begin{equation}
\begin{aligned}
&U_i(X_t,\boldsymbol a_i',A_{-i})-U_i(X_t,\boldsymbol a_i,A_{-i})\\
&\quad=\Phi_P(X_t,\boldsymbol a_i',A_{-i})
                   -\Phi_P(X_t,\boldsymbol a_i,A_{-i}).
\end{aligned}
\label{eq:potential-identity}
\end{equation}
Thus a payoff-improving unilateral change also increases the common
potential \citep{monderer1996potential}. Maximizing this potential supplies
a mutually consistent reference for the state-conditioned planning game.
}

Acceleration preferences alone do not prevent a vehicle from advancing
into occupied space. We therefore constrain each sequence using a
curvature-dependent speed cap \(\bar v_i\) and an available stopping
distance \(s_i^{\rm stop}\ge0\). For each prediction step
\(k=0,\ldots,H_P-1\), the constraints are
\begin{equation}
\begin{aligned}
 -a_{\rm brake}&\le a_{i,k}\le a_{\rm drive},\\
 0&\le v_i+\Delta_P\sum_{\ell=0}^{k}a_{i,\ell}
                  \le\max(\bar v_i,v_i),\\
 v_i\Delta_P(k+1)
   &+\Delta_P^2\sum_{\ell=0}^{k}
       \big(k-\ell+\tfrac12\big)a_{i,\ell}
                  \le s_i^{\rm stop},
\end{aligned}
\label{eq:pg-feasible}
\end{equation}
where the positive constants \(a_{\rm brake}\) and \(a_{\rm drive}\)
specify actuator limits.
The three inequalities bound acceleration, predicted speed, and travel
before a stopping point, respectively. The travel bound uses trapezoidal
integration and is omitted when no stop is required. Stopping distances
account for footprint obstacles and conflict occupancy; unoccupied
conflicts use estimated entry-time precedence. The speed bound permits
deceleration from a current speed above the curvature cap without making
the initial state infeasible. Together, these constraints define
\(\mathcal F(X_t)\).

The optimizer maximizes the potential over this feasible set and retains
the first acceleration of each sequence:
\begin{equation}
\begin{aligned}
 A_t^\star&\in\arg\max_{A\in\mathcal F(X_t)}\Phi_P(X_t,A),\\
 a_t^{\rm ref}&=\mathcal P_{X_t}
             \big((a_{i,0}^\star)_{i\in\mathcal C_t}\big),\\
 u_t^{\rm ref}&=(a_t^{\rm ref},P_t),
\end{aligned}
\label{eq:reference}
\end{equation}
where \(a_{i,0}^\star\) is the first sequence entry and \(P_t\) collects
the committed CAV paths. The physical projection \(\mathcal P_{X_t}\)
applies braking-distance, curvature-dependent speed, and actuator limits
at the current state. These limits are shared by reference and learned
commands. Numerical fallback when the linear program is infeasible is
specified in \ref{app:scenario}.

{
Since the constraints in \eqref{eq:pg-feasible} define per-vehicle feasible
sequences \(\mathcal F_i(X_t)\) at the observed state, a potential maximizer
satisfies
\begin{equation}
U_i(X_t,A_t^\star)\ge
 U_i(X_t,\boldsymbol a_i,A_{-i,t}^\star),\quad
 \forall\boldsymbol a_i\in\mathcal F_i(X_t),\quad i\in\mathcal C_t.
\label{eq:reference-equilibrium}
\end{equation}
No participating CAV can improve its modeled payoff by changing its
sequence alone while the others retain theirs. This equilibrium property
concerns the fixed-path planning solution; the projection in
\eqref{eq:reference} determines the acceleration passed to execution.
}

\rev{The fixed-path formulation provides structured longitudinal coordination
but restricts the spatial choices available to cooperating vehicles.
Changing acceleration alters when a conflict is encountered, while retaining
the underlying path-overlap relationships. Adjusting local paths can instead
remove some overlaps or open a passage around an occupied region.
The residual policy therefore extends the equilibrium-based reference to
joint motion and path decisions, allowing cooperation to change both the
timing and geometry of the interaction.}

\subsection{Residual MARL Policy Learning}
\label{sec:learning}

The PG reference coordinates motion along existing paths; releasing a
spatial obstruction additionally requires deciding how those paths and
their approach times should change together. We parameterize these
decisions relative to PG through a longitudinal acceleration residual,
a conflict-approach timing residual, and a lateral path residual.
The first two modify the longitudinal request, while the third extends
the available path geometry. Conditioning the policy on the reference
allows it to learn coordinated departures from an existing decision
rather than construct the entire joint maneuver from scratch.

At each decision instant, the PG solution is included in the observations
before the shared actor generates residuals for the \rev{participating CAVs}.
The policy is a diagonal Gaussian in three raw action coordinates:
\begin{equation}
\begin{aligned}
 z_{i,t}&\sim\mathcal N\!\left(\mu_\theta(o_{i,t}),
                       \operatorname{diag}(\sigma_\theta^2)\right),\\
 z_{i,t}&=(z^a_{i,t},z^\tau_{i,t},z^y_{i,t}),\qquad
 b_{i,t}=\tanh z_{i,t},
\end{aligned}
\label{eq:actor}
\end{equation}
where \(\pi_\theta(z_{i,t}\mid o_{i,t})\) has shared mean network
\(\mu_\theta\) and learned, state-independent standard deviations
\(\sigma_\theta\). The observation \(o_{i,t}\) contains ego motion,
route information, PG acceleration, and nearby vehicle states. The
superscripts \(a\), \(\tau\), and \(y\) identify the three residual
channels. Applying \(\tanh\) gives bounded decoder inputs
\(b^a_{i,t}\), \(b^\tau_{i,t}\), and \(b^y_{i,t}\); deterministic
evaluation replaces the sample by \(\mu_\theta(o_{i,t})\).
The observation construction is given in \ref{app:implementation}.

The two longitudinal channels distinguish an immediate acceleration
correction from a correction tied to an approaching conflict. Let
\(d_{i,t}\) be the distance to the nearest eligible geometric conflict
station, and define \(v^\dagger_{i,t}=\max(v_{i,t},v_{\rm floor})\),
where \(v_{\rm floor}>0\) regularizes low speeds. The timing channel
shifts the nominal time to this station and converts the resulting speed
request into an acceleration correction:
\begin{equation}
\begin{aligned}
 \tau_{i,t}&=d_{i,t}/v^\dagger_{i,t},\\
 v^{\rm tar}_{i,t}&=
 \frac{d_{i,t}}{\max(\tau_{\min},\tau_{i,t}+k_\tau b^\tau_{i,t})},\\
 \delta a^\tau_{i,t}
 &=\operatorname{clip}\!\left(
 \frac{v^{\rm tar}_{i,t}-v^\dagger_{i,t}}{T_v},
       -a_\tau,a_\tau\right),\\
 a_{i,t}^{\rm cand}
 &=a_{i,t}^{\rm ref}+k_a b^a_{i,t}+\delta a^\tau_{i,t},
\end{aligned}
\label{eq:longitudinal}
\end{equation}
where \(\tau_{\min}>0\) bounds the requested time from below,
\(k_\tau\) scales its shift, and \(T_v>0\) converts speed error into
acceleration. The constants \(a_\tau\) and \(k_a\) bound the timing
contribution and scale the direct acceleration residual, respectively.
Positive \(b^\tau_{i,t}\) requests a later nominal arrival; negative
values request an earlier one. If no station is eligible,
\(\delta a^\tau_{i,t}=0\). The timing channel therefore expresses an
approach-time preference through the longitudinal acceleration request.
Both longitudinal channels contribute to \(a_{i,t}^{\rm cand}\), whose
execution follows the physical limits of the selected path.

A longitudinal correction cannot bypass a vehicle that blocks the current
path. The lateral channel therefore requests a signed local offset
\(e_{i,t}=e_{\max}b^y_{i,t}\), where \(e_{\max}\) bounds its magnitude.
The decoder converts this scalar into a smooth departure from and return
to the current path. With distance \(s\) measured from the present
position along the remaining path \(\mathbf p_i(s)\) and unit normal
\(\mathbf n_i(s)\), the candidate is
\begin{equation}
\begin{aligned}
 \mathbf p_i^{\rm cand}(s)&=\mathbf p_i(s)
                  +e_{i,t}b_i^{\rm path}(s)\mathbf n_i(s),\\
 b_i^{\rm path}(s)&=
 h\!\left(\frac{s}{L_i}\right)-h\!\left(\frac{s-L_i}{L_r}\right),\\
 h(x)&=10\bar x^3-15\bar x^4+6\bar x^5,\\
 \bar x&=\operatorname{clip}(x,0,1),
\end{aligned}
\label{eq:lateral}
\end{equation}
where \(L_i\) and \(L_r\) specify the offset transition and return
distances. The smoothstep function \(h\) has zero first and second
derivatives at its endpoints, avoiding an abrupt offset change at the
beginning and end of the maneuver. The destination is unchanged.
Candidate paths are checked against road boundaries and steering limits;
curvature-dependent speed limits restrict lateral acceleration during
execution. The lateral residual specifies a path adjustment whose geometry
and speed profile determine the executed steering and lateral motion.
Equations~\eqref{eq:longitudinal} and~\eqref{eq:lateral} define the
decoder \(u_t^{\rm cand}=\mathcal D(X_t,u_t^{\rm ref},z_t)\).
Accepted paths become part of the next state, allowing subsequent PG
decisions to coordinate motion along them. Replanning conditions and
handling of infeasible paths are specified in \ref{app:scenario}.

An offset that frees one vehicle can obstruct another, so the usefulness
of a correction depends on the resulting joint motion. To provide this
interaction-level guidance, we predict the consequences of candidate
corrections and the reference along their respective paths.
A finite-horizon rollout
\(\mathcal R_H(X_t,u)\) applies joint command \(u\) for one decision
interval and recomputes PG on the resulting paths thereafter. Let
\(X_H(u)\) denote its terminal state after at most \(H\) decisions.
The rollout utility is
\begin{equation}
\begin{aligned}
 X_H(u)&=\mathcal R_H(X_t,u),\qquad J_H(u)=Q(X_H(u)),\\
 Q(X)&=-\frac1N\sum_{i=1}^N\ell_i(X)
          +\frac{\lambda_q}{N}\sum_{i=1}^N f_i(X),
\end{aligned}
\label{eq:rollout-utility}
\end{equation}
where \(\ell_i(X)\ge0\) is the remaining length of vehicle \(i\)'s
current path, \(f_i(X)\) indicates destination clearance, and
\(\lambda_q>0\) weights completion. Remaining distance accounts for
the extra travel introduced by a detour; completion rewards passage out
of the interaction. The rollout also records binary indicators
\(c_{\mathcal C}(u)\), \(c_{\mathcal N}(u)\), \(b(u)\), and \(d(u)\)
for a CAV-involved collision, any collision, a road-boundary violation,
and a dynamic-limit violation anywhere along the prediction.
Utility and feasibility are therefore evaluated separately. The same
comparison is used for execution in Section~\ref{sec:selector}.

Feasible, utility-improving corrections sampled at PG-visited states
provide action targets for the residual policy, with an auxiliary loss
\(\mathcal L_{\rm sup}\) described in \ref{app:targets}.
Shared-actor MAPPO then learns from their consequences over complete
interactions using a centralized critic \citep{yu2022surprising}.
During training and evaluation alike, the decoder constructs the joint
proposal and the selector either applies it or retains PG. Learning thus
accounts for the actual \rev{execution process}, including its fallback decisions.

To reward useful passage rather than motion alone, the team reward combines
remaining-distance reduction, completion, and penalties for unresolved or
unsafe traffic. Let \(\bar\ell_t=N^{-1}\sum_i\ell_i(X_t)\) be the
mean remaining path length before decoding the current proposal. Then
\begin{equation}
\begin{aligned}
 r_t={}&w_\ell(\bar\ell_t-\bar\ell_{t+1})
       +w_f\frac{n^{\rm new}_t}{N}
       -w_u\frac{n^{\rm unfin}_{t+1}}{N}\\
       &-w_c C^{\mathcal C}_{t+1}
       -w_b\mathbf1_{\{B_{t+1}\lor D_{t+1}\}}
       -w_r(1-g_t),
\end{aligned}
\label{eq:reward}
\end{equation}
where \(n^{\rm new}_t\) counts newly cleared vehicles and
\(n^{\rm unfin}_{t+1}\) counts those still unfinished. The flags
\(C^{\mathcal C}_{t+1}\), \(B_{t+1}\), and \(D_{t+1}\) record realized
CAV-involved collision, boundary violation, and dynamic-limit violation,
respectively. The acceptance indicator \(g_t\) is 1 when the learned
candidate is selected; \(\mathbf1_{\{\cdot\}}\) denotes an indicator
function. Positive weights balance the terms.
The first five terms instantiate the traffic objective \(R\) in
\eqref{eq:task}; the final term additionally discourages proposals that
are repeatedly rejected. Because \(\bar\ell_t\) is measured before a
new path is inserted, the progress term includes its added distance.
Unlike the finite-horizon utility \(J_H\), this reward is accumulated
over realized policy interactions and assigns an ongoing cost to waiting.

Joint outcomes provide a shared learning signal even though each CAV
generates its own residual. From the joint information \(O_t\) available
to the decision system, we construct the feature matrix
\(\mathbf O_t\in\mathbb R^{N_{\max}\times d_o}\), where \(N_{\max}\)
is the maximum vehicle count and \(d_o\) the observation dimension.
Each row is a vehicle observation \(o_{i,t}\), with zero padding for
absent vehicles. This finite feature representation is the neural-network
input; PG and the rollout comparison additionally use the committed-path
state in \(X_t\). The centralized critic \(V_\phi(\mathbf O_t)\)
estimates the team return. Its standardized generalized advantage estimate
\(\tilde A_t\), defined in \ref{app:updates}, is shared across
active CAVs. MAPPO retains an individual likelihood ratio for each CAV
and minimizes the masked actor loss
\begin{equation}
\begin{aligned}
 \rho_{i,t}(\theta)&=
 \frac{\pi_\theta(z_{i,t}|o_{i,t})}
      {\pi_{\theta_{\rm old}}(z_{i,t}|o_{i,t})},\\
 \bar\rho_{i,t}&=\operatorname{clip}(\rho_{i,t},1-\epsilon,1+\epsilon),\\
 \mathcal L_{\rm actor}
 &=\rev{-\left\langle\min(\rho_{i,t}\tilde A_t,
             \bar\rho_{i,t}\tilde A_t)\right\rangle_m
       -\beta\langle\mathcal H(\pi_\theta)\rangle_m
       +\lambda_{\rm sup}\mathcal L_{\rm sup}},
\end{aligned}
\label{eq:ppo}
\end{equation}
where \(\langle\cdot\rangle_m\) averages over active CAV--time
pairs, with mask \(m_{i,t}=1\) for a \rev{participating CAV}. The clipping
threshold \(\epsilon\), entropy weight \(\beta\), and supervised-loss
weight \(\lambda_{\rm sup}\) regulate the update; \(\mathcal H\)
denotes entropy. The parameters \(\theta_{\rm old}\) define the policy
that collected the batch. The critic is fitted to team-return targets
as detailed in \ref{app:updates}.

Policy updates use the likelihood of the sampled proposal, including
when it leads to fallback: the proposal is the actor's action, while the
decoder and selector determine its realized transition and reward.
The shared critic connects individual residuals to collective passage.
With this \rev{learning objective} established, the next question is how to supply
experience of the interactions in which useful corrections are needed.

\subsection{Hard-Case Curriculum Training}
\label{sec:curriculum}

\rev{Fixed-path coordination already resolves many routine encounters,
so repeatedly training on them offers limited opportunity to improve
cooperative passage. The larger opportunity lies in interactions where
longitudinal adjustment leaves vehicles mutually blocked and joint motion
and path changes are needed. To direct training toward these cases, we
evaluate a candidate set \(\mathcal S_{\rm pool}\) using PG alone and
retain its poorly resolved scenes in a fixed difficult subset
\(\mathcal S_{\rm hard}\). The reference's episode outcomes therefore
determine where additional learning experience is allocated.}

Difficulty must distinguish unsafe passage, incomplete passage, and passage
that succeeds only after substantial waiting. For a training scene \(s\),
let \(C(s)\) indicate a CAV-involved collision under PG and let \(S(s)=1\)
indicate that all vehicles clear without any collision or invalid motion.
For successful as well as unsuccessful scenes, define delay by
\begin{equation}
\begin{aligned}
 D(s)&=\frac1{N_s}\sum_{i=1}^{N_s}
           \left[\tilde t_i-\frac{\ell_i^0}{v_{\max}}\right]_+,\\
 \tilde t_i&=
 \begin{cases}
 t_i^{\rm clear},&\text{if vehicle }i\text{ clears},\\
 T_{\max},&\text{otherwise}
 \end{cases},
\end{aligned}
\label{eq:delay}
\end{equation}
where \(N_s\) is the vehicle count, \(\ell_i^0\) the initial route length,
\(t_i^{\rm clear}\) the elapsed clearance time, \(T_{\max}\) the episode
limit, and \([x]_+=\max(x,0)\). The speed limit \(v_{\max}\) defines
the nominal travel time. Unfinished vehicles contribute their waiting up
to the deadline, so delay remains defined for incomplete passage.
The three outcomes rank scenes in descending lexicographic order:
\begin{equation}
 \mathcal S_{\rm hard}
 =\operatorname{TopK}^{\rm lex}_{s\in\mathcal S_{\rm pool}}
       \big(C(s),\,1-S(s),\,D(s)\big),
\label{eq:hardpool}
\end{equation}
where \(K\) is the difficult-subset size. Collisions take precedence over
other failures, and delay separates scenes with the same collision and
success status. Thus safety, completion, and waiting enter the ranking
in that order, identifying interactions requiring safer or more effective
coordination from the reference's episode outcomes.

To retain scene diversity while emphasizing these interactions, training
mixes the difficult subset with freshly generated scenes. At policy
update \(j\), the sampling distribution is
\begin{equation}
 p_j=\eta_j\,\operatorname{Unif}(\mathcal S_{\rm hard})
                    +(1-\eta_j)p_{\rm fresh},
\label{eq:curriculum}
\end{equation}
where \(\operatorname{Unif}\) samples uniformly from the difficult
subset and \(p_{\rm fresh}\) generates new vehicle counts, routes,
initial states, and CAV assignments. The share \(\eta_j\) increases
through training, giving unresolved interactions greater emphasis after
the policy has acquired initial correction behavior. The schedule is
specified in \ref{app:updates}.
The curriculum changes the allocation of experience while retaining the
same action representation and reward. This gives poorly resolved
interactions greater influence on learning; whether a resulting correction
is useful in the current state is determined separately at execution.

\rev{Algorithm~\ref{alg:coopdistill-training} summarizes this training process:
PG outcomes identify difficult scenes, the actor receives correction-level
guidance, and sampled interactions update the actor and centralized critic
through MAPPO.}

\subsection{Comparative Rollout Selector}
\label{sec:selector}

Training equips the policy to propose corrections, but their value depends
on the vehicles and occupied space present at the decision instant.
A path adjustment can change the conflict relationships used to construct
PG. \rev{A maneuver that helps one vehicle may consequently obstruct another
or introduce a collision, even when it was favored during training.}
We therefore compare the complete learned proposal with the current
reference using the finite-horizon rollout in \eqref{eq:rollout-utility},
and execute it only when the comparison supports doing so.

Both rollouts start from identical copies of \(X_t\) and use the same
HDV response model and physical limits. Each branch first applies its own
joint command, then recomputes PG along the resulting paths. HDV responses
are recomputed as CAV motion changes, rather than held to a trajectory
predicted independently of the proposal. The acceptance indicator is
\begin{equation}
\begin{aligned}
 g_t=1\ \Longleftrightarrow\quad
 &c_{\mathcal C}(u_t^{\rm cand})=0,\\
 &b(u_t^{\rm cand})=d(u_t^{\rm cand})=0,\\
 &c_{\mathcal N}(u_t^{\rm cand})
       \le c_{\mathcal N}(u_t^{\rm ref}),\\
 &J_H(u_t^{\rm cand})\ge J_H(u_t^{\rm ref})-\epsilon_J,
\end{aligned}
\label{eq:acceptance}
\end{equation}
where \(\epsilon_J\ge0\) is a numerical utility tolerance. The first
two conditions reject predicted CAV-involved collisions and invalid
motion. The third prevents accepting a proposal that introduces an
any-vehicle collision absent from the reference prediction. Among
candidates satisfying these checks, the last condition retains only
those whose rollout utility is at least that of PG within the tolerance.
This separates motion feasibility from the comparison of predicted
traffic progress.

Selection acts on the joint command, preserving the coordinated vehicle
actions evaluated together in each rollout. The command passed to the
environment is therefore
\begin{equation}
\begin{aligned}
 u_t^{\rm exec}&=
 \begin{cases}
 u_t^{\rm cand},&g_t=1,\\
 u_t^{\rm ref},&g_t=0,
 \end{cases}\\
 X_{t+1}&=F(X_t,u_t^{\rm exec};\xi_t).
\end{aligned}
\label{eq:execution}
\end{equation}

Accepted paths are committed before the physical limits are applied.
Otherwise, vehicles retain their current paths and execute the PG request.
The same \rev{selection process} is used during training, so its acceptance and rejection
outcomes enter the policy's experience through \eqref{eq:reward}.
Acceptance is reassessed at every decision under the current rollout
model and prediction horizon. Fallback uses PG on the committed paths,
including any earlier accepted correction. The reference therefore
continues to coordinate traffic after a path change, while the policy
can propose further improvements as the interaction evolves.
\rev{Algorithm~\ref{alg:coopdistill-execution} summarizes online decision generation,
rollout comparison, and execution of the selected joint command.}

{\SetAlgoNoEnd
\begin{algorithm}[t]
\caption{Training of \method}
\label{alg:coopdistill-training}
\KwIn{Candidate training set $\mathcal S_{\rm pool}$, fresh scenes drawn from $p_{\rm fresh}$, and update count $U$}
\KwOut{Trained residual actor $\pi_\theta$ and centralized critic $V_\phi$}
Initialize actor parameters $\theta$ and critic parameters $\phi$\;
\Comment{Identify difficult training scenes}
Roll out every $s\in\mathcal S_{\rm pool}$ to termination using PG only\;
Construct $\mathcal S_{\rm hard}$ from collision, success, and delay using \eqref{eq:hardpool}\;
\Comment{Learn reference-conditioned residuals}
Fit $\theta$ to rollout-evaluated corrections using \ref{app:targets}\;
\For{$j\leftarrow0$ \KwTo $U-1$}{
  Set $\theta_{\rm old}\leftarrow\theta$\;
  \ForEach{rollout slot}{
    Draw scene $s\sim p_j$ according to \eqref{eq:curriculum}; reset $X_0$ and $t\leftarrow0$\;
    \While{the episode is not terminal}{
      Compute $u_t^{\rm ref}$ with \eqref{eq:reference} and construct $\mathbf O_t$\;
      Sample each active CAV's $z_{i,t}$ with \eqref{eq:actor} and decode the joint candidate $u_t^{\rm cand}$\;
      Compare both rollouts, evaluate $g_t$ with \eqref{eq:acceptance}, and execute \eqref{eq:execution}\;
      Store $\mathbf O_t,z_t,r_t$, action log probabilities, and the active-CAV mask\;
      $t\leftarrow t+1$\;
    }
  }
  Compute \eqref{eq:gae} and update $(\theta,\phi)$ with \eqref{eq:ppo} and \eqref{eq:critic}\;
}
\end{algorithm}}
{\SetAlgoNoEnd
\begin{algorithm}[t]
\caption{Online execution of \method}
\label{alg:coopdistill-execution}

\KwIn{Initial traffic state $X_0$ and trained residual actor $\pi_\theta$}
\KwOut{Executed joint decisions and terminal traffic state}
$t\leftarrow0$\;
\While{the episode is not terminal}{
  \Comment{Construct the reference and learned proposal}
  Compute $u_t^{\rm ref}$ with \eqref{eq:reference}\;
  Set $z_{i,t}=\mu_\theta(o_{i,t})$ for each active CAV and decode $u_t^{\rm cand}$ with \eqref{eq:longitudinal}--\eqref{eq:lateral}\;
  \Comment{Compare predicted consequences from the same state}
  Roll out $u_t^{\rm ref}$ and $u_t^{\rm cand}$ from cloned copies of $X_t$ using $\mathcal R_H$\;
  Evaluate the acceptance indicator $g_t$ with \eqref{eq:acceptance}\;
  \Comment{Commit the candidate or fall back to PG}
  \eIf{$g_t=1$}{
    Commit candidate paths and set $u_t^{\rm exec}\leftarrow u_t^{\rm cand}$\;
  }{
    Retain current paths and set $u_t^{\rm exec}\leftarrow u_t^{\rm ref}$\;
  }
  Apply physical limits and advance to $X_{t+1}$ as in \eqref{eq:execution}\;
  $t\leftarrow t+1$\;
}
\end{algorithm}}
\endgroup

\section{Experiments}
\label{sec:experiments}

\rev{Experiments evaluate \method in a merging area, a roundabout, and an
unsignalized intersection. Comparisons across CAV penetration levels
assess safety and passage efficiency, followed by component ablations
and trajectory analyses of how learned corrections release blocked traffic.}

\subsection{Experimental Setup}

\textbf{\rev{Simulation environment:}}
\rev{The three environments contain six to twelve vehicles per scene, with
road-specific vehicle counts and geometry detailed in \ref{app:scenario}.
Routes, initial positions, speeds, and CAV assignments vary across scenes.
HDVs use heterogeneous IDM-like car-following and conflict-yielding behavior,
while CAVs receive cooperative decisions. Simulation and decision intervals
are \(0.1\,\mathrm{s}\) and \(0.5\,\mathrm{s}\), respectively.}

\textbf{\rev{Evaluation protocol:}}
\rev{Each scenario is tested at target CAV penetrations of 25\%, 50\%,
75\%, and 100\%, using 120 common cases per level for all methods.
Scenario generation and checkpoint protocols are specified in
\ref{app:evaluation-protocol}.}

\rev{The six baselines are potential-game coordination
(PG; \citealp{fang2026adaptive}), first-come-first-served reservation
(FCFS; \citealp{dresner2008multiagent,malcolm2025lanefree}), auction reservation
(Auction; \citealp{carlino2013auction,suriyarachchi2022gameopt}), Nash-bargaining MPC
(NMPC; \citealp{arevalo2023bargaining}), continuous-action MAPPO
(\citealp{yu2022surprising}), and a discrete-action multi-agent deep Q-network
(MADQN; \citealp{fan2026joint}). The cited decision principles are implemented
with shared vehicle dynamics, physical constraints, and test scenes.}

\rev{During evaluation, four metrics are used. \emph{Success} is the fraction of episodes in
which all vehicles clear within the horizon without collision, boundary
violation, or dynamic-limit violation. \emph{Collision} counts episodes
with a CAV-involved collision; HDV--HDV collisions are excluded from this
metric but invalidate success. \emph{Cleared} is the fraction of vehicles
reaching their destinations. \emph{Delay} follows \eqref{eq:delay}, assigning
the episode deadline to vehicles that do not clear. Success and collision
are averaged over episodes, while clearance and delay are pooled over vehicles.}

\subsection{Main Results}

\textbf{Overall performance:}
\rev{Table~\ref{tab:main-comparison} reports performance pooled across the four
penetration groups. \method attains the highest episode success in the
intersection and roundabout and shares the highest success in merging.
Relative to PG, success increases by 2.1, 27.1, and 13.1 percentage points,
respectively, while delay decreases in every setting. The joint improvement
in completion and waiting shows that the learned corrections extend the
reference's cooperative capability, rather than achieving progress by
accepting more CAV-involved collisions. The larger gains in the roundabout
and merging area are consistent with the value of adjusting local paths
when fixed-path longitudinal coordination leaves traffic blocked.}

\begin{table}[t]
\centering
\caption{Performance across three cooperative driving scenarios.}
\label{tab:main-comparison}
\small
\setlength{\tabcolsep}{2.5pt}
\begin{tabular*}{\textwidth}{@{\extracolsep{\fill}}llrrrrrr|r@{}}
\toprule
Environment & Metric & PG & FCFS & Auction & NMPC & MAPPO & MADQN & \method\\
\midrule
\multirow{4}{*}{Merging area} & Success (\%) & 86.9 & \cellcolor{TableBest!75}\textbf{100.0} & \cellcolor{TableBest!75}\textbf{100.0} & 90.2 & \cellcolor{TableBest!75}\textbf{100.0} & \cellcolor{TableSecond!75}96.0 & \cellcolor{TableBest!75}\textbf{100.0}\\
 & Collision (\%) & \cellcolor{TableBest!75}\textbf{0.0} & \cellcolor{TableBest!75}\textbf{0.0} & \cellcolor{TableBest!75}\textbf{0.0} & \cellcolor{TableBest!75}\textbf{0.0} & \cellcolor{TableBest!75}\textbf{0.0} & \cellcolor{TableSecond!75}1.0 & \cellcolor{TableBest!75}\textbf{0.0}\\
 & Cleared (\%) & 93.3 & \cellcolor{TableBest!75}\textbf{100.0} & \cellcolor{TableBest!75}\textbf{100.0} & 95.7 & \cellcolor{TableBest!75}\textbf{100.0} & \cellcolor{TableSecond!75}97.2 & \cellcolor{TableBest!75}\textbf{100.0}\\
 & Delay (s) & 13.7 & 15.9 & 15.9 & 16.8 & 12.9 & \cellcolor{TableSecond!75}12.5 & \cellcolor{TableBest!75}\textbf{10.3}\\
\midrule
\multirow{4}{*}{Roundabout} & Success (\%) & 71.7 & 47.9 & 66.9 & \cellcolor{TableSecond!75}98.3 & 97.1 & 98.1 & \cellcolor{TableBest!75}\textbf{98.8}\\
 & Collision (\%) & \cellcolor{TableBest!75}\textbf{0.0} & \cellcolor{TableBest!75}\textbf{0.0} & \cellcolor{TableBest!75}\textbf{0.0} & \cellcolor{TableBest!75}\textbf{0.0} & \cellcolor{TableBest!75}\textbf{0.0} & \cellcolor{TableSecond!75}0.2 & \cellcolor{TableBest!75}\textbf{0.0}\\
 & Cleared (\%) & 86.3 & 80.8 & 89.1 & \cellcolor{TableSecond!75}99.4 & \cellcolor{TableBest!75}\textbf{99.7} & \cellcolor{TableSecond!75}99.4 & \cellcolor{TableBest!75}\textbf{99.7}\\
 & Delay (s) & 30.0 & 33.1 & 30.2 & 22.0 & 23.5 & \cellcolor{TableBest!75}\textbf{20.0} & \cellcolor{TableSecond!75}21.8\\
\midrule
\multirow{4}{*}{Intersection} & Success (\%) & \cellcolor{TableSecond!75}86.9 & 60.0 & 78.5 & 77.1 & 78.5 & 74.8 & \cellcolor{TableBest!75}\textbf{89.0}\\
 & Collision (\%) & \cellcolor{TableBest!75}\textbf{0.0} & \cellcolor{TableBest!75}\textbf{0.0} & \cellcolor{TableBest!75}\textbf{0.0} & \cellcolor{TableBest!75}\textbf{0.0} & 0.8 & \cellcolor{TableSecond!75}0.6 & \cellcolor{TableBest!75}\textbf{0.0}\\
 & Cleared (\%) & \cellcolor{TableSecond!75}92.6 & 81.5 & 87.9 & 88.0 & 87.6 & 83.0 & \cellcolor{TableBest!75}\textbf{93.7}\\
 & Delay (s) & \cellcolor{TableSecond!75}11.5 & 16.1 & 13.6 & 13.0 & 11.8 & 14.0 & \cellcolor{TableBest!75}\textbf{10.6}\\
\bottomrule
\end{tabular*}
\par\smallskip
{\footnotesize Blue (bold): best; green: second best. Ties are determined at the displayed precision.\par}
\end{table}

\begin{figure}[t]
\centering
\includegraphics[width=\textwidth]{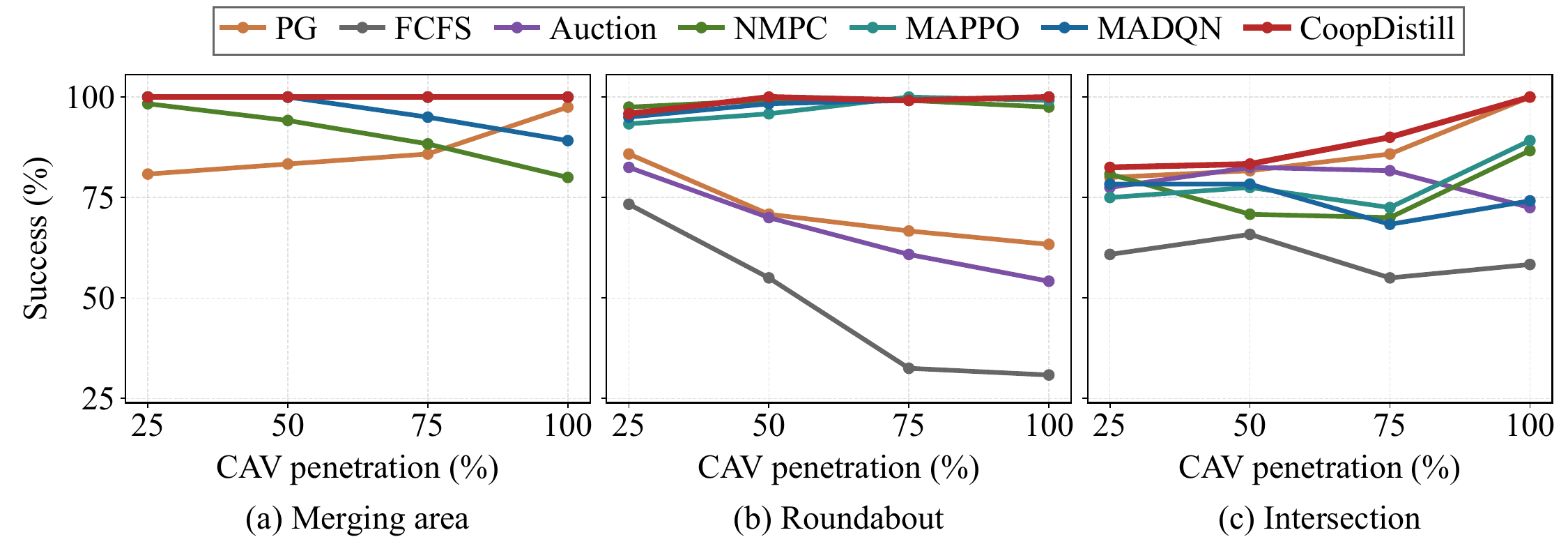}
\caption{Episode success across CAV penetration levels in three scenarios.}
\label{fig:penetration-success}
\end{figure}

\rev{Figure~\ref{fig:penetration-success} \revexp{further separates} these results by CAV
penetration. \method clears every merging episode successfully and
maintains roundabout success above 95\%. Roundabout PG success falls
as more vehicles are assigned to cooperative control, whereas \method
maintains high completion. This contrast shows that additional cooperative
vehicles alone are insufficient: their actions must also resolve the spatial
interactions created by lane use and entry--exit movements. At the
intersection, \method improves success over PG in all three mixed-traffic
groups, with the largest gain at 75\%; both reach full success at 100\%.}

\begin{figure}[t]
\centering
\includegraphics[width=\textwidth]{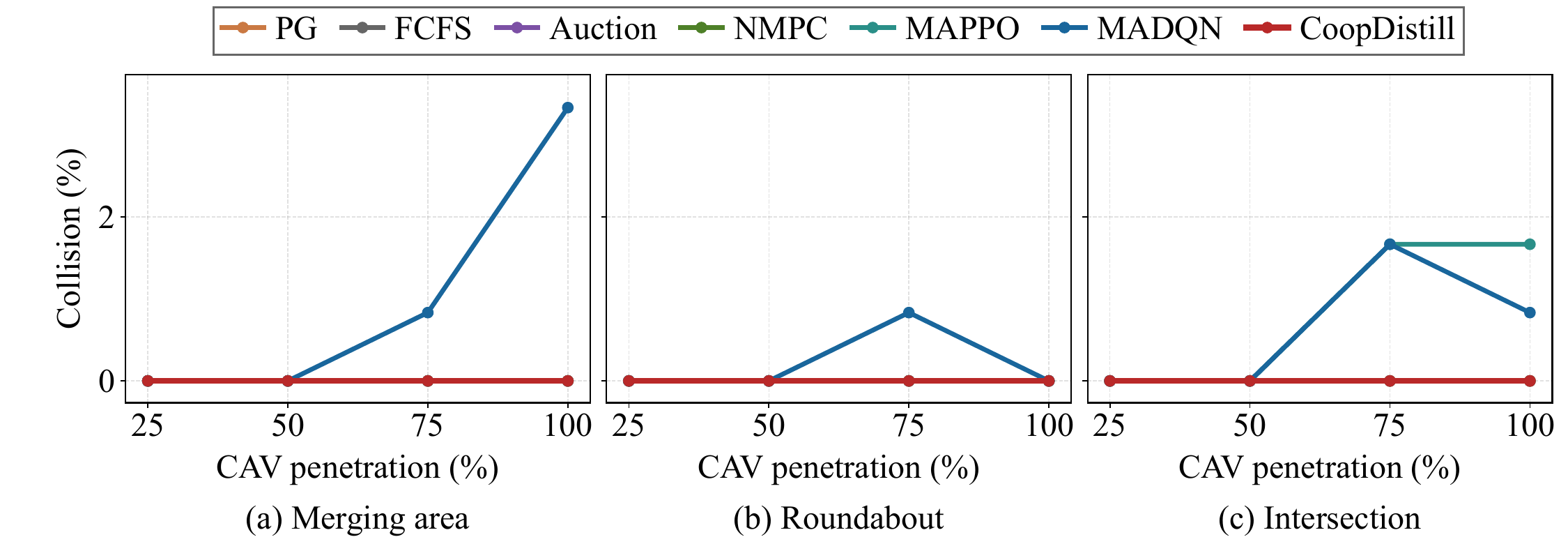}
\caption{CAV-involved collision across penetration levels. Coincident zero-collision curves overlap.}
\label{fig:penetration-collision}
\end{figure}

\textbf{Safety:}
\revexp{Figure~\ref{fig:penetration-collision} records no CAV-involved collisions
for \method at any tested penetration. PG, FCFS, Auction, and NMPC also
record zero, whereas MAPPO incurs a 0.8\% collision rate at the intersection
and MADQN incurs collisions in all three environments, reaching 1.0\%
in merging. These collisions occur despite the shared physical execution
limits: learning a reward-maximizing policy does not itself enforce
collision avoidance between interacting vehicles. This distinction
motivates the comparative rollout selector, which checks the joint
consequences of learned proposals against PG before execution.
\method thereby retains the collision performance of the structured
methods while completing more interactions than PG.}

\begin{figure}[t]
\centering
\includegraphics[width=\textwidth]{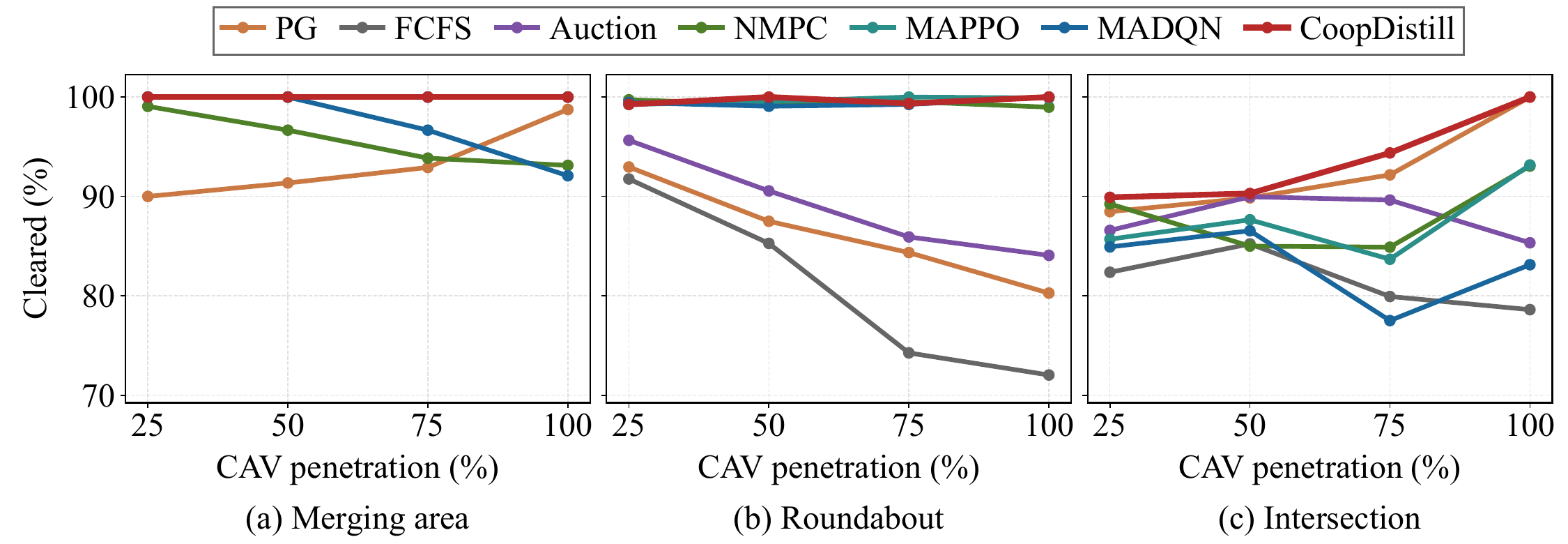}
\caption{Vehicle clearance across CAV penetration levels in three scenarios.}
\label{fig:penetration-cleared}
\end{figure}

\begin{figure}[t]
\centering
\includegraphics[width=\textwidth]{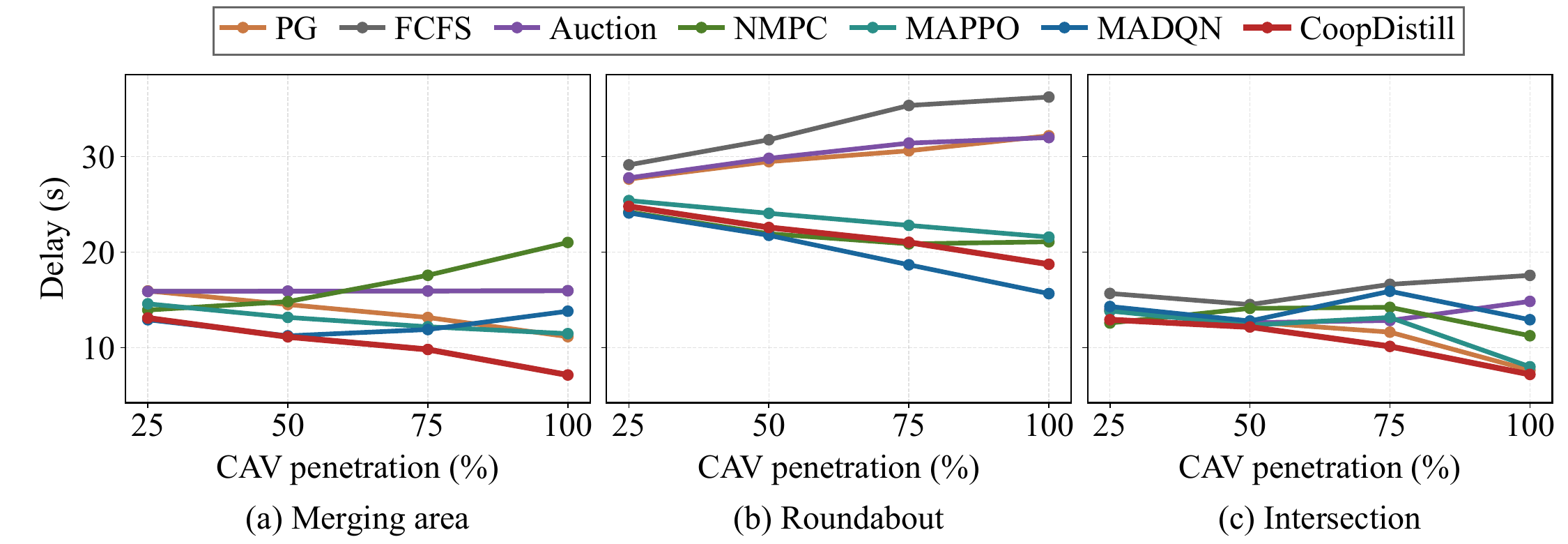}
\caption{Mean vehicle delay across CAV penetration levels. Unfinished vehicles are assigned the episode deadline.}
\label{fig:penetration-delay}
\end{figure}

\textbf{Efficiency:}
\rev{Figures~\ref{fig:penetration-cleared} and~\ref{fig:penetration-delay}
show how much traffic is released and the waiting incurred during passage.
Compared with PG, \method reduces mean delay by 24.8\% in merging,
27.3\% in the roundabout, and 7.3\% at the intersection. In merging,
FCFS, Auction, MAPPO, and \method all clear every vehicle, but \method
achieves the lowest pooled delay. This comparison isolates an efficiency
benefit beyond completing more episodes: coordinated motion and path
adjustments allow vehicles to use available gaps with less waiting.}

\rev{As CAV penetration increases, \method's delay decreases across all
three settings while clearance remains high or improves. At full
intersection penetration, PG and \method both clear every vehicle,
but \method still reduces delay by 5.7\%. Its local trajectory corrections
therefore improve passage even when the reference already completes the
interaction. \method has the lowest intersection delay at 50\%, 75\%, and
100\%; at 25\%, NMPC has lower delay while \method achieves higher success
and clearance. In the roundabout, MADQN achieves the lowest mean delay,
whereas \method has higher success and shares the highest clearance.
The complete comparison thus concerns completion, safety, and waiting
together, rather than a ranking based on a single efficiency measure.}

\begin{figure}[t]
\centering
\includegraphics[width=\textwidth]{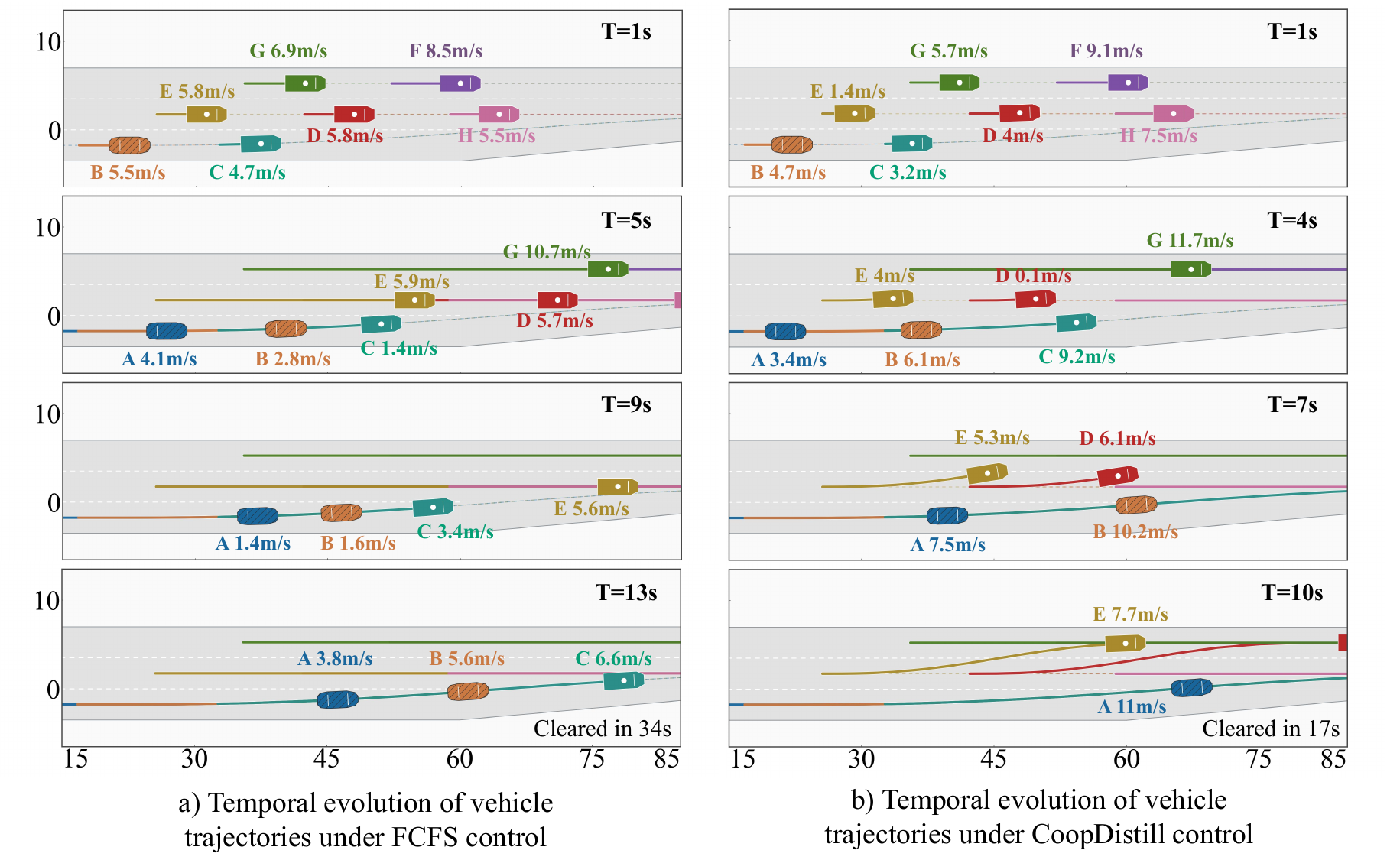}
\caption{Cooperative merging under (a) FCFS and (b) \method.}
\label{fig:merge-fcfs-case}
\end{figure}

\textbf{Representative interactions:}
Figure~\ref{fig:merge-fcfs-case} compares FCFS and \method from identical
initial states in an eight-vehicle merging scene. Video demonstrations of
the case studies are available on the project website\footnote{\url{https://fangshiyuu.github.io/CoopDistill/#demonstrations}}.
Colors distinguish
vehicles, and the annotations report simulation time and instantaneous
speed. Thin dashed lines show reference paths, whereas thick solid lines
show traveled trajectories. Angular bodies with a white roof marker
denote CAVs; rounded, hatched bodies denote HDVs.
Under FCFS, vehicles
following the ramp remain slow while the adjacent mainline stream
advances. \method adjusts both progress and local paths: vehicles D and E
move toward the other mainline lane, releasing space for the merging
vehicles. The snapshots show the merging vehicles advancing sooner under
\method, with scene clearance taking approximately 17~s compared with
34~s under FCFS. Both runs finish without collision. This example
illustrates how coordinated spatial adjustments can
reduce waiting even when both methods ultimately achieve full clearance.

\begin{figure}[t]
\centering
\includegraphics[width=\textwidth]{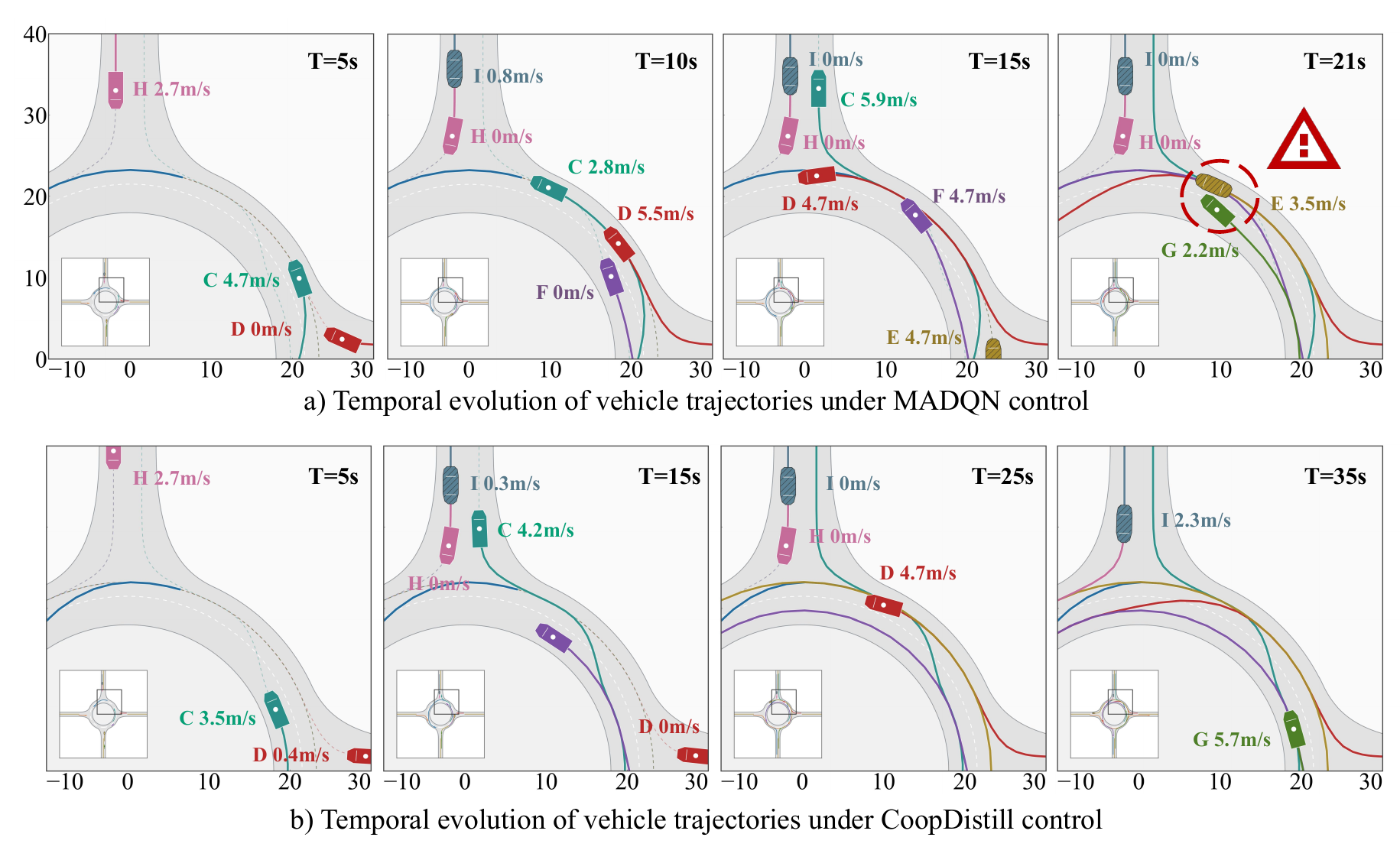}
\caption{Local roundabout interactions under (a) MADQN and (b) \method.}
\label{fig:ring-madqn-case}
\end{figure}

\revexp{Figure~\ref{fig:ring-madqn-case} compares MADQN and \method from
identical initial states in a nine-vehicle roundabout scenario, focusing
on an entry--exit interaction area. Under MADQN, several vehicles advance
rapidly through the circulating roadway, but the converging movements of
vehicles E and G lead to a collision at approximately 21~s. \method instead
coordinates local path adjustments with longitudinal progression.
Vehicle D waits while C passes through the shared area, then resumes
motion as space becomes available; G subsequently progresses through
the same region without collision. This sequence illustrates that effective
cooperation requires coordinating both where vehicles travel and when
they proceed, rather than maximizing individual forward motion. Within
\method, residual actions provide maneuver flexibility, while PG
coordination and comparative rollout selection organize their execution
around the surrounding traffic.}

\FloatBarrier

\subsection{Ablation and Mechanism Analysis}

\revexp{The ablations examine how residual MARL policy learning and hard-case
curriculum training contribute to cooperative passage, with the comparative
rollout selector retained during execution.}
Table~\ref{tab:ablation} reports equal-weight averages over three fixed
validation sets, with matched training budgets and the same cases for
all variants within each scenario.
The variant \emph{w/o residual} replaces residual parameterization with
absolute longitudinal acceleration and lateral path requests and removes
the associated correction-target fitting and auxiliary loss. It still receives
PG reference information. The \emph{\revexp{w/o curriculum}} variant retains
residual actions and continuing correction-target supervision while disabling
hard-case sampling, whereas
\emph{w/o residual \& curriculum} removes both components.
\revexp{\method combines residual MARL policy learning with hard-case curriculum training.}
The decoder and comparative selector remain enabled in all four learned
variants. Validation-set sizes and checkpoint selection are specified in
\ref{app:evaluation-protocol}.

\begin{table}[!htbp]
\centering
\caption{Component comparison on three fixed validation sets.}
\label{tab:ablation}
\small
\begin{tabular}{@{}lcccc@{}}
\toprule
Variant & Success (\%) & Collision (\%) & Cleared (\%) & Delay (s)\\
\midrule
PG & 80.7 & 0.0 & 86.9 & 20.1\\
w/o residual \& curriculum & 91.7 & 0.0 & 93.7 & 16.3\\
\revexp{w/o curriculum} & 93.8 & 0.0 & 94.5 & 15.8\\
w/o residual & 88.9 & 0.0 & 91.7 & 16.5\\
\midrule
CoopDistill & 94.4 & 0.0 & 95.4 & 15.2\\
\bottomrule
\end{tabular}
\end{table}

\FloatBarrier

\begin{figure}[!htbp]
\centering
\includegraphics[width=\linewidth]{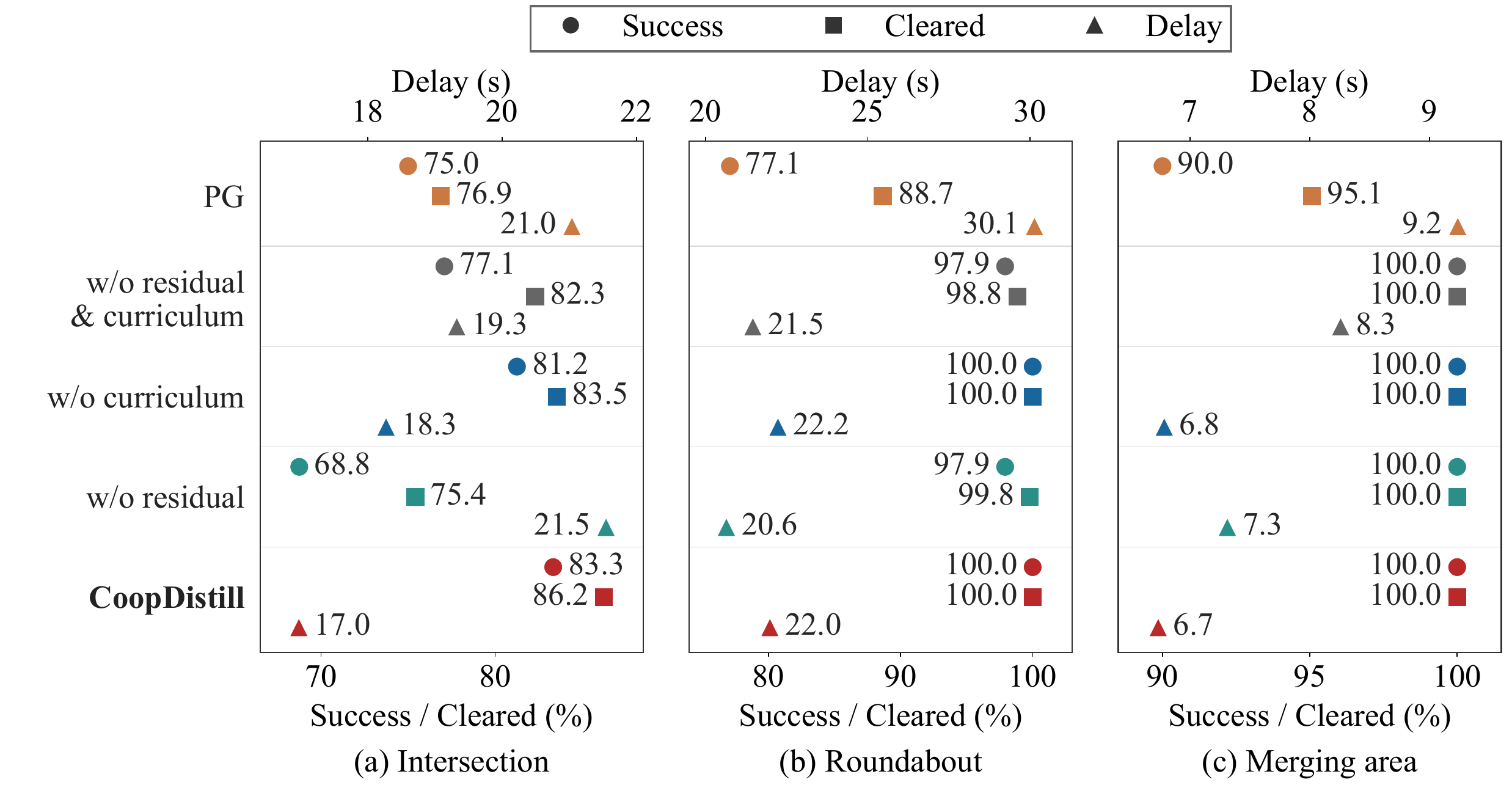}
\caption{Ablation performance by scenario on fixed validation sets. Circles and squares indicate success and clearance; triangles indicate delay.}
\label{fig:validation-ablation-scenes}
\end{figure}
\FloatBarrier

\textbf{Overall performance:}
\revexp{\method achieves the highest mean success and clearance and the lowest
mean delay, with no observed CAV-involved collisions. Relative to PG,
mean success improves by 13.8 percentage points and delay decreases by
24.4\%. Compared with removing both residual MARL and curriculum,
the full design improves success by 2.8 percentage points and reduces
delay by 1.1~s. The w/o curriculum variant approaches the full model,
but \method retains a 0.7-percentage-point success advantage and a
0.5~s delay reduction.}

\revexp{Figure~\ref{fig:validation-ablation-scenes} presents the ablation
results separately for the three scenarios. At the intersection, the
complete framework improves completion and delay: success reaches
83.3\%, compared with 81.3\% for w/o curriculum and 68.8\% for w/o residual.
The full design thus releases more intersection traffic with less waiting.
In merging, all learned variants attain full clearance, and the lowest
delays are obtained by the variants retaining residual MARL.
Residual corrections therefore remain useful after completion has
already saturated. In the roundabout, both \method and w/o curriculum
complete every case, with their difference appearing in delay.
Across scenarios, the full framework combines improved intersection
completion with efficient merging and reliable roundabout passage.
All variants record zero CAV-involved collisions, so the plotted
differences concern completion and waiting.}

\begin{figure}[H]
\centering
\includegraphics[width=0.65\linewidth]{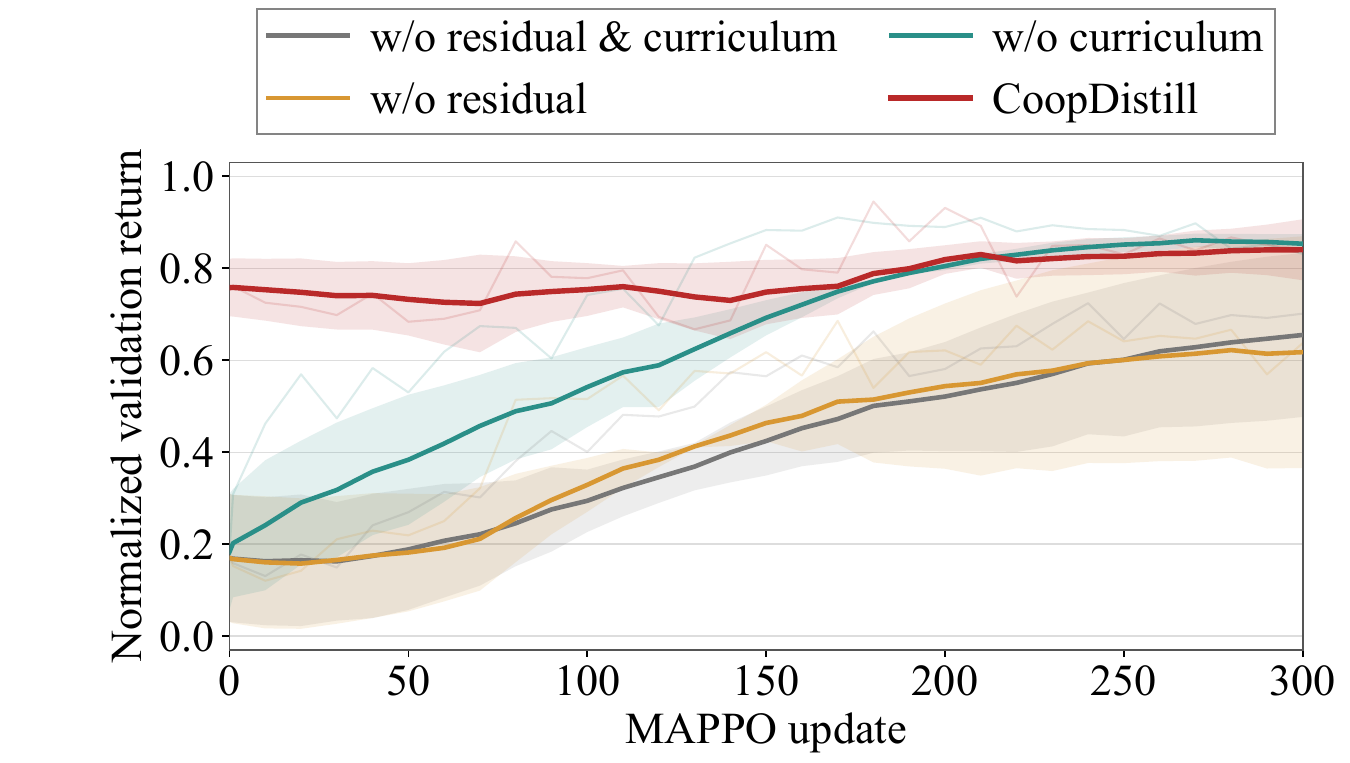}
\caption{Learning curves on fixed validation sets. Lines show the smoothed mean normalized return across three scenarios; shading indicates one standard deviation across scenarios.}
\label{fig:normalized-learning-return}
\end{figure}

\textbf{Training returns:}
Figure~\ref{fig:normalized-learning-return} compares the four ablation
variants using discounted returns on fixed validation scenes. Returns
are normalized within each scenario and averaged equally across
scenarios with identical exponential smoothing; shading indicates
between-scenario standard deviation. \method maintains high returns
from the outset, while the other variants improve from lower starting
points. Although residual learning without curriculum eventually
approaches \method, its returns are substantially lower early in
training. With curriculum alone, but without residual learning, returns
rise more slowly and remain below those of the residual variants.
Concentrating experience on difficult interactions is therefore most
effective when coupled with a policy that can build on the PG reference.

The higher starting return of \method comes from fitting its residual
policy to feasible, rollout-improving corrections around PG before
MAPPO updates. The policy thus begins with useful correction behavior
instead of having to discover joint motion and path adjustments from
scratch. Subsequent residual exploration and hard-case training
develop this behavior through environmental interaction.

\FloatBarrier

\begin{figure}[!htbp]
\centering
\includegraphics[width=\linewidth]{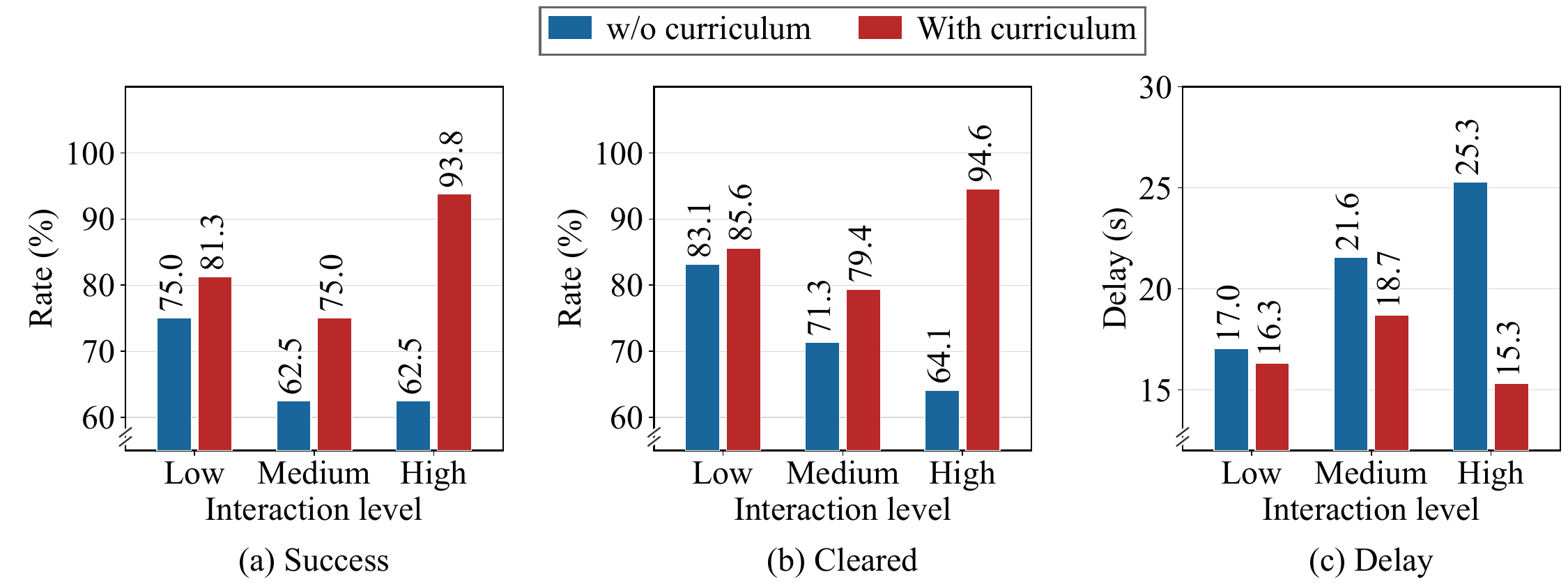}
\caption{Effect of curriculum across intersection interaction levels: (a) success; (b) clearance; (c) delay. Vertical axes are truncated.}
\label{fig:curriculum-intersection}
\end{figure}

\begin{samepage}
\textbf{Curriculum across interaction levels:}
To identify which interactions benefit from targeted experience,
Fig.~\ref{fig:curriculum-intersection} compares training with and without
curriculum on the fixed intersection validation set.
\revexp{An initial interaction graph connects vehicles whose reference-path
footprint occupancies overlap within an arrival-time window. Its mean
degree measures the number of interacting neighbors per vehicle.
The lower and upper tertiles across distinct initial layouts define
low, medium, and high interaction levels, independently of policy outcomes.}
Both variants retain the same residual learning and
execution mechanisms and are compared at a matched training budget
(\ref{app:evaluation-protocol}).
\par
\end{samepage}

\FloatBarrier

\revexp{The largest curriculum benefit occurs under high interaction,
where several vehicles must adjust their motion around overlapping
paths. Success increases from 62.5\% to 93.8\%, a gain of 31.3 percentage
points, while clearance rises from 64.1\% to 94.6\% and delay falls
from 25.3 to 15.3~s, a 39.5\% reduction. The corresponding success
gains are 6.3 and 12.5 percentage points under low and medium interaction.
The much larger improvement in strongly interacting traffic directly
supports the purpose of hard-case curriculum training: concentrating
experience on unresolved coordination rather than repeatedly sampling
routine passage. With the maneuver space and execution mechanism held
fixed, this experience allocation enables the policy to release more
vehicles from mutual obstruction with less waiting.}

\begin{table}[t]
\centering
\caption{Selector behavior and paired success of the full ablation model.}
\label{tab:selector-analysis}
\small
\begin{tabular}{@{}lr@{}}
\toprule
Statistic & Result\\
\midrule
MARL proposal accepted & 69.2\%\\
PG fallback invoked & 30.8\%\\
PG failure $\rightarrow$ Full success & 14.3\%\\
PG success $\rightarrow$ Full failure & 1.0\%\\
Exact paired label-swap test & $p<0.001$\\
\bottomrule
\end{tabular}
\end{table}

\textbf{Selector behavior and statistical significance:}
\revexp{Table~\ref{tab:selector-analysis} describes the execution of the full
ablation model on the formal test sets. Learned joint proposals are
accepted for 69.2\% of decisions on average, while PG is used for the
remaining 30.8\%. The learned policy therefore contributes actions
frequently, and the reference remains an active alternative when the
candidate fails the rollout comparison. This is the intended division
of responsibility: residual learning expands the available maneuvers,
and selection determines when their predicted consequences justify execution.}

Paired outcomes distinguish recovered interactions from new failures.
Averaged across scenarios, Full converts 14.3\% of episodes from
PG failure to success, while the reverse transition occurs in 1.0\%.
Both transition rates use all paired episodes as the denominator.
The overall success difference is significant under a layout-clustered
paired test ($p<0.001$).

\begin{figure}[t]
\centering
\includegraphics[width=\linewidth]{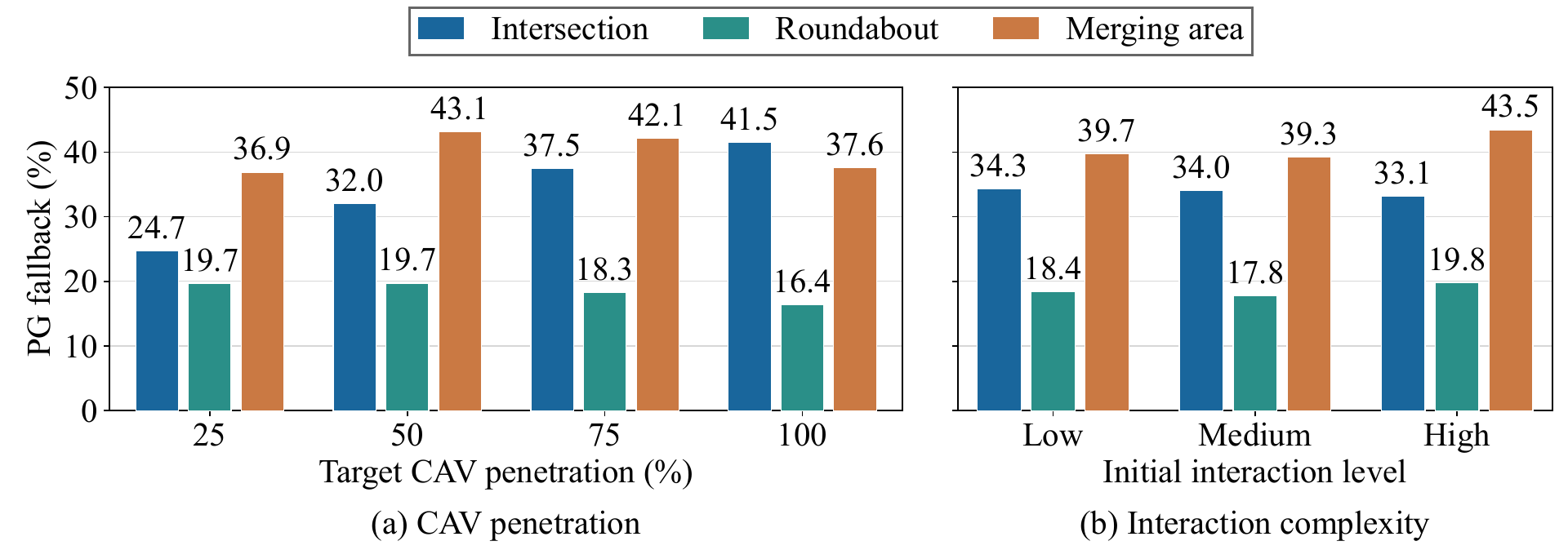}
\caption{PG fallback across (a) CAV penetration and (b) initial interaction levels.}
\label{fig:fallback-diagnostics}
\end{figure}

\revexp{Figure~\ref{fig:fallback-diagnostics} further breaks down the fallback
decisions in Table~\ref{tab:selector-analysis} by CAV penetration and initial
interaction level. Each bar averages episode-level fallback fractions
within its group. Case-weighted aggregation of either panel recovers
the overall fallback rate of 30.8\% in Table~\ref{tab:selector-analysis}.}

\revexp{Fallback varies with road geometry and CAV participation. It increases
with penetration at the intersection, decreases slightly overall in the
roundabout, and peaks at intermediate penetration in merging. Across
interaction levels, its fraction remains comparatively stable at the
intersection and roundabout, while highly interacting merging scenes
use PG more often. The selector therefore responds to the particular
joint proposal and available reference, rather than applying a fixed
switching fraction for difficult traffic. A higher fallback fraction
means that PG is executed more frequently; difficulty is measured
separately from the initial interaction structure.}

\textbf{\revexp{Representative interactions:}}
\begingroup
\revexp{Figure~\ref{fig:CoopDistill_case7} compares PG and \method from identical
initial conditions in a seven-vehicle intersection encounter. Under PG, vehicles
enter the intersection and yield to occupied conflict regions until
mutual waiting prevents further passage. Longitudinal acceleration
adjustments change when vehicles approach these regions but retain
the path overlaps responsible for the obstruction. The interaction
therefore requires a spatial change, rather than further adjustment
of fixed-path approach times alone.}

\begin{figure*}[t]
\centering
\includegraphics[width=\textwidth]{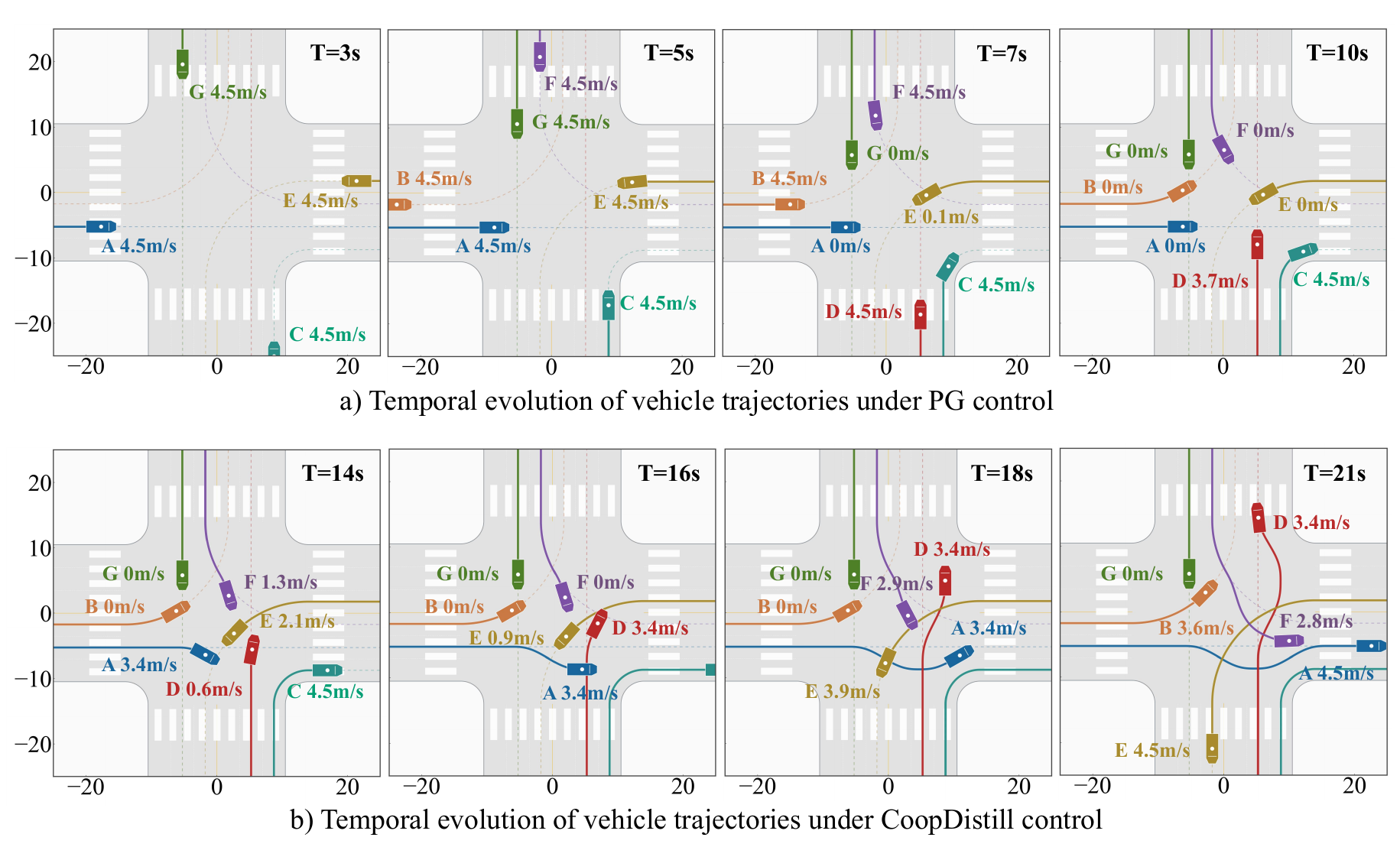}
\caption{Intersection passage under (a) PG and (b) \method.}
\label{fig:CoopDistill_case7}
\end{figure*}

\revexp{Residual MARL supplies this additional maneuver flexibility:
local lateral path corrections create room around blocked vehicles,
while longitudinal acceleration and conflict-approach timing residuals
coordinate progress along the adjusted paths. Comparative rollout
selection evaluates the resulting joint maneuver against PG before
execution, and the reference continues to coordinate motion along
accepted paths. As one vehicle advances, it releases occupied space
for others, allowing the obstruction to dissipate without changing
their destinations. The comparison connects the learned expansion
of the maneuver space and reference-based execution to the observed
gains in cooperative passage.}

\endgroup

Taken together, the experiments address the three challenges motivating
\method. The return curves show that reference-guided residual learning
provides high-return behavior early in training. The interaction-level
comparison shows that targeted experience improves completion and
reduces waiting most in strongly interacting scenes. Rollout diagnostics
and trajectory comparisons illustrate how learned proposals and PG
fallback jointly support execution. These findings connect maneuver
exploration, training experience allocation, and execution-time selection
to improved cooperative passage.

\FloatBarrier
\section{Conclusion}

\rev{This paper presented \method, an equilibrium-guided multi-agent policy
distillation framework for cooperative driving. A PG reference coordinates
fixed-path longitudinal motion, residual MARL extends its maneuver space,
hard-case training emphasizes unresolved interactions, and rollout comparison
selects the executed joint decision. Experiments in merging, roundabout,
and intersection scenarios yield three main findings.}

\begin{enumerate}
\item \rev{The framework improves cooperative passage relative to PG in all
three road settings. Success increases by 13.1, 27.1, and 2.1 percentage
points in merging, roundabout, and intersection scenarios, respectively,
while delay decreases by 7.3--27.3\%, with no observed CAV-involved collisions.}
\item \rev{The full design attains the highest mean success and clearance
and lowest mean delay in the component comparison. The interaction-level
comparison shows that curriculum produces its largest gains in strongly
interacting intersection scenes, where concentrating training experience
improves completion and reduces waiting.}
\item \rev{Learned corrections and the PG reference both contribute to execution.
The selector accepts learned proposals for 69.2\% of joint decisions on
average. Trajectory comparisons show how local path adjustments change
obstruction relationships and allow vehicles to clear interactions that
remain blocked under fixed-path coordination.}
\end{enumerate}

Future work will extend this framework to sustained traffic with
continuous arrivals and interacting bottlenecks, and investigate
rollout decisions under uncertain driver responses and communication
delays. These extensions will connect the coordination of individual
encounters to longer-term traffic operations.

\FloatBarrier
\appendix
\begingroup

\section{Implementation and Training Details}
\label{app:implementation}

This appendix provides the observation representation, action-target
construction, and policy updates underlying residual learning, followed
by the curriculum schedule. Numerical network and training settings below
describe the intersection model in the main comparison. Scene capacity
and geometry follow the configuration used for each experiment; the
ablation evaluation protocol is specified in
\ref{app:evaluation-protocol}.

\subsection{Observation and Network Parameterization}

The policy requires local interaction information and the current reference,
whereas the critic requires the joint traffic state available to learning.
Both use the padded observation matrix defined in
Section~\ref{sec:learning}. In the intersection implementation, each
vehicle occupies one 44-entry normalized row.
The first 16 entries encode
the vehicle's position, heading sine and cosine, speed, acceleration, remaining
route distance, PG acceleration, current lateral offset, CAV indicator, exit-gate
coordinates, accumulated waiting time, completion flag, and left- and right-turn
indicators. The next 24 entries contain six features for each of the four nearest
unfinished vehicles: relative position, world-frame velocity, CAV indicator, and
slot-occupancy flag. The final four entries encode elapsed time, vehicle count,
active-vehicle status, and active-CAV status. Neighbor entries outside the
\(80\,\mathrm{m}\) observation radius and unused vehicle rows are zero. Actor
losses are masked to unfinished CAVs, whereas the critic receives all eight
rows and can therefore condition on CAV and HDV motion jointly.

The shared actor is a multilayer perceptron with two 128-unit tanh layers and a
three-dimensional output mean. Its state-independent diagonal log standard
deviation is initialized to \(-0.8\) and learned jointly with the mean network.
The centralized critic flattens the \(8\times44\) observation matrix and uses
192- and 128-unit tanh layers before a scalar value output. Stochastic raw
Gaussian samples are used during training, while deterministic evaluation uses
the actor mean. In both cases, the same decoder and comparative rollout
selector are applied. The stored raw Gaussian sample is used to calculate
policy likelihoods before bounding and applying physical limits. HDV and
finished-vehicle rows remain available to the critic but do not contribute
to the actor loss; absent rows are zero.

\subsection{Action-Target Construction}
\label{app:targets}

The actor's initial targets provide examples of corrections with favorable
predicted consequences. They use the decoder and rollout utility already
defined in Section~\ref{sec:learning}. States are collected from independent
PG episodes, without executing the sampled corrections in those episodes.
At a sampled state \(X\), draw joint residuals
\(z^{(k)}\sim\mathcal N(0,\sigma_s^2 I)\) and decode
\(u^{(k)}=\mathcal D(X,u^{\rm ref},z^{(k)})\),
where \(k\) indexes candidates, \(\sigma_s\) is their sampling standard
deviation, and the identity matrix \(I\) spans all active-CAV action components.
Initialize \(J_{\rm best}=J_H(u^{\rm ref})\) and \(z^\star=0\), where
\(J_H\) and the violation indicators are defined in
\eqref{eq:rollout-utility} and its accompanying text.
For each candidate in sampled order, update
\begin{equation}
\begin{aligned}
 (z^\star,J_{\rm best})&\leftarrow(z^{(k)},J_H(u^{(k)}))
                   \quad\text{if}\\
 c_{\mathcal N}(u^{(k)})&=b(u^{(k)})=d(u^{(k)})=0,\\
 J_H(u^{(k)})&>J_{\rm best}+\epsilon_s.
\end{aligned}
\label{eq:search}
\end{equation}
Unlike online acceptance, target selection requires a strict utility
improvement and excludes any predicted collision, including HDV--HDV
collisions. A candidate replaces the incumbent only when it improves
utility by more than \(\epsilon_s>0\). Its raw sampled residual is
stored as the label, avoiding an inverse mapping from the executed path
and acceleration. Improving labels receive weight \(w=1\); if none
is found, the zero label receives the smaller weight \(w_0<1\).
The resulting observation--target data set
\(\mathcal D_{\rm sup}=\{(o_i,z_i^\star,w)\}\) defines
\begin{equation}
 \mathcal L_{\rm sup}(\theta)=
 \frac{1}{|\mathcal D_{\rm sup}|}
 \sum_{(o,z^\star,w)\in\mathcal D_{\rm sup}}
       \frac{w}{3}\|\mu_\theta(o)-z^\star\|_2^2.
\label{eq:supervised}
\end{equation}
This loss initializes the actor and remains as a small auxiliary term in
\eqref{eq:ppo}; it does not replace the return-based MAPPO objective.
The target-collection episodes are separate from
\(\mathcal S_{\rm pool}\), whose PG outcomes are used only to rank
training scenes. Thus correction targets and curriculum difficulty have
different roles: one supplies action-level guidance, while the other
determines the distribution of subsequent interaction experience.

\subsection{Policy Updates and Scene Sampling}
\label{app:updates}

MAPPO estimates the consequence of a joint proposal from the realized
reward sequence, including any fallback chosen by the selector. For an
episode containing \(T_e\) decisions, generalized advantage estimation
uses the value parameters \(\phi_{\rm old}\) before the update:
\begin{equation}
\begin{aligned}
 \delta_t&=r_t+\gamma V_{\phi_{\rm old}}(\mathbf O_{t+1})
                         -V_{\phi_{\rm old}}(\mathbf O_t),\\
 \hat A_t&=\sum_{q=0}^{T_e-t-1}(\gamma\lambda)^q\delta_{t+q},\\
 \hat Y_t&=\hat A_t+V_{\phi_{\rm old}}(\mathbf O_t),
\end{aligned}
\label{eq:gae}
\end{equation}
where \(\lambda\) controls advantage smoothing, and
\(V_{\phi_{\rm old}}(\mathbf O_{T_e})=0\) because the deadline belongs to the
finite-horizon task. Standardizing \(\hat A_t\) over time steps containing
an active CAV gives \(\tilde A_t\). The actor uses a separate likelihood
ratio for each active CAV with this shared advantage. The critic minimizes
\begin{equation}
 \mathcal L_{\rm critic}(\phi)
   =\frac12\left\langle
       \big(V_\phi(\mathbf O_t)-\hat Y_t\big)^2
     \right\rangle_t,
\label{eq:critic}
\end{equation}
where the average covers all collected time steps.

The scene sampling share in \eqref{eq:curriculum} increases linearly:
\begin{equation}
 \eta_j=\eta_0+(\eta_1-\eta_0)\frac{j}{\max(U-1,1)},
 \quad j=0,\ldots,U-1.
\label{eq:curriculum-schedule}
\end{equation}
The initial and final shares are \(\eta_0\) and \(\eta_1\), and \(U\)
is the number of updates. Each episode draws independently from the fixed
difficult subset with probability \(\eta_j\), otherwise from fresh scenes.
The difficult subset is ranked once and does not change during training.
No scene-difficulty label is supplied to the deployed actor.

\subsection{Training Settings}

Table~\ref{tab:appendix-training} lists the intersection training settings.
The actor and critic are optimized separately with Adam. Generalized
advantage estimates are normalized over valid rollout time steps, and gradients
are clipped independently for the two networks. The reported intersection model uses
training seed 23; its checkpoint at update 150 is selected by the lexicographic
validation criterion of safety, success, delay, and speed on a split disjoint
from the reported test scenes.

\begin{table}[t]
\centering
\caption{Intersection training settings.}
\label{tab:appendix-training}
\small
\begin{tabular}{@{}lr@{}}
\toprule
Setting & Value\\
\midrule
Policy updates / episodes per update & 240 / 12\\
PPO epochs / minibatch size & 4 / 256\\
Actor / critic learning rate & \(3\!\times\!10^{-4}\) / \(10^{-3}\)\\
Discount \(\gamma\) / GAE \(\lambda\) & 0.99 / 0.95\\
PPO clipping threshold \(\epsilon\) & 0.20\\
Entropy coefficient \(\beta\) & 0.005\\
Supervised-loss weight \(\lambda_{\rm sup}\) & 0.02\\
Actor / critic gradient norm limit & 0.5 / 1.0\\
Candidate training set / difficult subset & 120 / 32\\
Initial / final hard-case share & 0.25 / 0.50\\
\bottomrule
\end{tabular}
\end{table}

The action targets in \eqref{eq:search} are collected from 12 independent
PG episodes, using at most eight states from each episode.
Eight joint residual candidates are drawn
per sampled state from \(\mathcal N(0,0.85^2I)\). The strict improvement
tolerance in \eqref{eq:search} is \(10^{-4}\); unsuccessful states retain the
zero label with weight \(w_0=0.2\). The actor is initialized for 16 epochs, and
the resulting fixed data set supplies the supervised loss in subsequent
MAPPO updates. The reward weights in \eqref{eq:reward} are
\(w_\ell=0.1\), \(w_f=4\), \(w_u=0.05\), \(w_c=20\),
\(w_b=10\), and \(w_r=0.03\).

\section{Scenario and Evaluation Details}
\label{app:scenario}

This appendix specifies scenario generation, physical execution, and metric
aggregation for Section~\ref{sec:experiments}.

\subsection{Scene Generation}

All vehicles are present at the initial simulation time and follow legal
right-hand-traffic routes. For intersection training, each origin is
sampled uniformly, and left,
straight, and right movements are sampled with probabilities 0.30, 0.45, and
0.25, respectively. Training scenes contain four to eight vehicles. Dense
scenes place vehicles \(23\)--\(40\,\mathrm{m}\) from their entry, with initial
speeds of \(2.5\)--\(4.0\,\mathrm{m/s}\); ordinary scenes use
\(28\)--\(75\,\mathrm{m}\) and \(3.0\)--\(6.0\,\mathrm{m/s}\). Vehicles in the
same approach lane are separated by at least \(13\,\mathrm{m}\). HDV desired
speeds, time headways, and comfortable decelerations are sampled from
\([5.5,7.5]\,\mathrm{m/s}\), \([1.1,1.9]\,\mathrm{s}\), and
\([1.5,2.5]\,\mathrm{m/s^2}\), respectively. CAV identities are drawn without
replacement to realize the requested penetration as closely as the integer
vehicle count permits.

The original intersection hard set is constructed from 360 fixed candidate scenes in
three families: perturbations of a conflict-rich anchor, topology variations,
and independently sampled arrivals. A scene is retained only if it contains at
least three active approaches, five route-conflict pairs, and two conflict pairs
involving a CAV. Using PG outcomes only, selection is balanced across the three
families, targets 24 PG-failure scenes, and fills the remaining positions with
the highest-delay PG-success scenes. The resulting 120 layouts underpin
the 50\%, 75\%, and 100\% groups in the current comparison; the 25\%
group is generated and screened separately at that penetration.
No learned-policy result enters this selection. Both test cohorts are
separate from \(\mathcal S_{\rm pool}\) and \(\mathcal S_{\rm hard}\),
which supply training experience only. The merging test set fixes the
vehicle count at eight, while the roundabout balances eight to ten
vehicles across entry--exit weaving and heterogeneous-speed interactions.
\rev{The intersection test scenes contain six to eight vehicles.}

\subsection{Physical Limits and Rollout Settings}

The physical projection \(\mathcal P_{X_t}\) in \eqref{eq:reference}
limits acceleration using current braking distances, curvature-dependent
speed caps, and actuator bounds. These limits are recomputed at simulation
steps for both reference and learned commands. If the PG linear program
is infeasible or the solver fails, nominal car-following acceleration
subject to the same limits supplies the request. The travel constraints
in \eqref{eq:pg-feasible} use trapezoidal integration, whereas \(B_P\)
retains the potential's acceleration-to-travel approximation.
Each solve uses interaction coefficients fixed at the observed state;
subsequent decisions update them as vehicle motion and conflict geometry
change.

The analytic path in \eqref{eq:lateral} is sampled and reparameterized by
arc length for execution. Replanning requires a lateral path residual
above a deadband, sufficient remaining route length, and expiration of
the preceding path's hold time. If sampled vehicle footprints leave the
road or curvature exceeds the steering limit, the proposed path is
discarded. The current path is retained, and the longitudinal correction
can still undergo rollout comparison. Curvature-dependent speed limits
restrict lateral acceleration along the selected path.

A zero lateral path residual retains the current path. Zero longitudinal
acceleration and timing residuals recover the PG request when no timing
station is eligible or when \(\tau_{i,t}\ge\tau_{\min}\).
If the nominal time is below \(\tau_{\min}\), the denominator bound in
\eqref{eq:longitudinal} can instead add braking even at zero timing
residual. The zero training target therefore passes through the same
timing bound and physical limits as every other proposal. Once a path
is accepted, subsequent PG fallback operates along that committed path.

Table~\ref{tab:appendix-simulation} reports the intersection execution
settings. A PG prediction contains eight
\(0.5\,\mathrm{s}\) stages, whereas the comparative rollout selector simulates six
decisions, or \(3\,\mathrm{s}\), from each cloned current state. The rollout
utility in \eqref{eq:rollout-utility} uses \(\lambda_q=10\), and the online
utility tolerance in \eqref{eq:acceptance} is \(10^{-6}\). Collision checks use
oriented rectangular footprints at five substeps within every simulation step
to reduce missed overlaps at crossings.
The selector applies the PG request when the candidate fails acceptance,
including when both rollouts predict a violation; the shared physical
limits still apply. Its comparison is local to the current state and
prediction horizon, and subsequent decisions use the paths already committed.

\begin{table}[t]
\centering
\caption{Intersection simulation, decoder, and rollout settings.}
\label{tab:appendix-simulation}
\small
\begin{tabular}{@{}lr@{}}
\toprule
Setting & Value\\
\midrule
Vehicle length / width / wheelbase & 4.6 / 1.9 / 2.8 m\\
Maximum steering angle & \(46^{\circ}\)\\
Simulation / decision interval & 0.1 / 0.5 s\\
Episode horizon / speed limit & 60 s / 8 m/s\\
Acceleration range & \([-4,2]\,\mathrm{m/s^2}\)\\
PG / selector prediction horizon & 4 / 3 s\\
Maximum path offset & 3 m\\
Offset transition / return distance & 18 / 18 m\\
Maximum lateral acceleration & \(2.2\,\mathrm{m/s^2}\)\\
Timing-station eligibility window & \((1,30)\) m\\
Longitudinal acceleration residual scale & \(1.5\,\mathrm{m/s^2}\)\\
Conflict-approach timing shift bound & \(1.0\,\mathrm{s}\)\\
Timing-induced acceleration bound & \(0.8\,\mathrm{m/s^2}\)\\
\bottomrule
\end{tabular}
\end{table}

\subsection{Evaluation Sets and Checkpoints}
\label{app:evaluation-protocol}

The main comparison, learning ablations, and execution diagnostics use
distinct evaluation protocols. Table~\ref{tab:main-comparison} uses
validation-selected policies on the formal test sets, with 120 cases
per scenario and penetration. Each learned checkpoint is held fixed
across the four penetration groups.

Table~\ref{tab:ablation} and Fig.~\ref{fig:validation-ablation-scenes}
use fixed validation sets containing 48 intersection, 48 roundabout,
and 40 merging cases. The ablation runs share training seed 37 and
12 RL episodes per update. They are compared at update 200
(2,400 RL episodes), selected retrospectively from the archived
validation curves as a common budget.
The \revexp{w/o curriculum} variant in Table~\ref{tab:ablation} omits the preliminary
correction-target fitting stage and retains its auxiliary loss during MAPPO.
The curriculum-only comparison in Fig.~\ref{fig:curriculum-intersection}
retains the same preliminary fitting stage in both variants and compares
them at update 180 (2,160 episodes).
It partitions the same intersection validation set into three groups
of 16 cases according to initial reference-path occupancy and
arrival-time proximity. These displays report fixed-budget validation
performance, separately from the formal-test comparison.

Table~\ref{tab:selector-analysis} and
Fig.~\ref{fig:fallback-diagnostics} use validation-selected Full
checkpoints from the ablation runs on the formal test sets.
Acceptance and fallback count joint proposals while at least one CAV
remains unfinished. Rates are calculated within each episode, averaged
over episodes within each scenario, and then averaged equally across
scenarios. Interaction-level thresholds are fixed within each scenario
from initial path occupancy and arrival-time proximity, independently
of policy outcomes. The paired success analysis uses a two-sided exact
label-swap test, grouping all penetration outcomes of each physical
layout together. The combined test pools these layout-level groups
across scenarios, including the separate low-penetration intersection
layouts, and evaluates the fixed trained policies.

\subsection{Penetration Groups and Metric Aggregation}
\label{app:aggregation}

Table~\ref{tab:main-comparison} and
Figs.~\ref{fig:penetration-success}--\ref{fig:penetration-delay}
use the same four penetration groups: 25\%, 50\%, 75\%, and 100\%.
Each scenario contributes 120 episodes per group. All methods use
identical cases within a scenario and group; learned checkpoints are
fixed across the four groups. The roundabout uses the same physical
layouts across penetration levels with nested CAV assignments.
Intersection groups retain their archived PG-screened case selections.
Requested penetration is a scene-generation parameter: for \(N_s\)
vehicles and target fraction \(p\), the assigned CAV count is
\(\max(1,\min(N_s,\lfloor pN_s+0.5\rfloor))\).
The realized fraction can therefore differ from the requested value.

The averaging unit matters because scene sizes vary. For an evaluation
set \(\mathcal E\), let \(S_s\) and \(C_s\) be its episode success and
CAV-collision indicators, and let \(n_s^{\rm clear}\) count cleared
vehicles. The main and ablation table metrics are
\begin{equation}
\begin{aligned}
\mathrm{Success}&=\frac{\sum_{s\in\mathcal E}S_s}{|\mathcal E|},
&\mathrm{Collision}&=\frac{\sum_{s\in\mathcal E}C_s}{|\mathcal E|},\\
\mathrm{Cleared}&=\frac{\sum_{s\in\mathcal E}n_s^{\rm clear}}
                         {\sum_{s\in\mathcal E}N_s},
&\mathrm{Delay}&=\frac{\sum_{s\in\mathcal E}N_sD(s)}
                       {\sum_{s\in\mathcal E}N_s}.
\end{aligned}
\label{eq:aggregation}
\end{equation}
Proportions are expressed as percentages in the tables.
For Table~\ref{tab:ablation}, these metrics are computed within each
fixed validation set and then averaged equally over the three scenarios.
The main-comparison penetration curves apply the same within-scenario
aggregation separately to each group. Thus, Table~\ref{tab:main-comparison}
and its penetration figures differ only in whether the four groups are
pooled or displayed separately, not in their metric definitions.
Unfinished vehicles retain the deadline-assigned delay in both displays.

\endgroup

\section*{Acknowledgments}
This work was supported in part by the Interdisciplinary Disciplines Breakthrough Plan of the Ministry of Education of China (No.JYB2025XDXM123), the National Natural Science Foundation of China (52125208, 52302502).

\section*{Compliance with ethics guidelines}
Shiyu Fang, Sheng Zhao, Jiarui Zhang, Chengkai Xu, Xiaocong Zhao, Peng Hang, and Jian Sun declare that they have no conflict of interest or financial conflicts to disclose.

\bibliographystyle{elsarticle-harv}
\bibliography{references}

@article{fang2026adaptive,
  title={Toward Cooperative Driving in Mixed-Traffic: An Adaptive Potential Game-Based Approach with Field Test Verification},
  author={Fang, Shiyu and Zhao, Xiaocong and Liu, Xuekai and Hang, Peng and Wang, Jianqiang and Wang, Yunpeng and Sun, Jian},
  journal={Engineering},
  year={2026},
  doi={10.1016/j.eng.2026.03.022}
}

@article{malcolm2025lanefree,
  title={Lane-free intersection control for connected automated vehicles prioritizing vulnerable road users},
  author={Malcolm, Patrick and Bogenberger, Klaus},
  journal={Transportation Research Part C: Emerging Technologies},
  volume={170},
  pages={104918},
  year={2025},
  doi={10.1016/j.trc.2024.104918}
}

@inproceedings{suriyarachchi2022gameopt,
  title={{GAMEOPT}: Optimal Real-time Multi-Agent Planning and Control for Dynamic Intersections},
  author={Suriyarachchi, Nilesh and Chandra, Rohan and Baras, John S. and Manocha, Dinesh},
  booktitle={2022 IEEE 25th International Conference on Intelligent Transportation Systems (ITSC)},
  pages={2599--2606},
  year={2022},
  doi={10.1109/ITSC55140.2022.9921968}
}

@article{arevalo2023bargaining,
  title={Cooperative Cruise Control for Intelligent Connected Vehicles: A Bargaining Game Approach},
  author={Arevalo-Castiblanco, Miguel F. and Pachon, Jaime and Tellez-Castro, Duvan and Mojica-Nava, Eduardo},
  journal={Sustainability},
  volume={15},
  number={15},
  pages={11898},
  year={2023},
  doi={10.3390/su151511898}
}

@article{chen2026game,
  title={A Game-Theoretical Framework for Safe Decision Making and Control of Mixed Autonomy Vehicles},
  author={Chen, Mingyang and others},
  journal={IEEE Transactions on Intelligent Transportation Systems},
  volume={27},
  number={1},
  pages={1338--1351},
  year={2026},
  doi={10.1109/TITS.2025.3623584}
}

@article{fabiani2020multi,
  title={Multi-Vehicle Automated Driving as a Generalized Mixed-Integer Potential Game},
  author={Fabiani, Filippo and Grammatico, Sergio},
  journal={IEEE Transactions on Intelligent Transportation Systems},
  volume={21},
  number={3},
  pages={1064--1073},
  year={2020},
  doi={10.1109/TITS.2019.2901505}
}

@article{cao2025conflict,
  title={A Game Theoretic Decision-Making Framework With Conflict-Aware Nash Equilibrium Selection for Autonomous Vehicles at Uncontrolled Intersections},
  author={Cao, Yuxiao and Zeng, Xiangrui and Yin, Zhouping},
  journal={IEEE Transactions on Intelligent Transportation Systems},
  volume={26},
  number={1},
  pages={210--224},
  year={2025},
  doi={10.1109/TITS.2024.3483844}
}

@article{wang2023control,
  title={Learning to Control and Coordinate Mixed Traffic Through Robot Vehicles at Complex and Unsignalized Intersections},
  author={Wang, Dawei and Li, Weizi and Zhu, Lei and Pan, Jia},
  journal={arXiv preprint arXiv:2301.05294},
  year={2023}
}

@article{gorsane2022standardised,
  title={Towards a Standardised Performance Evaluation Protocol for Cooperative MARL},
  author={Gorsane, Rihab and Mahjoub, Omayma and de Kock, Ruan and Dubb, Roland and Singh, Siddarth and Pretorius, Arnu},
  journal={arXiv preprint arXiv:2209.10485},
  year={2022}
}

@article{dresner2008multiagent,
  title={A Multiagent Approach to Autonomous Intersection Management},
  author={Dresner, Kurt and Stone, Peter},
  journal={Journal of Artificial Intelligence Research},
  volume={31},
  pages={591--656},
  year={2008},
  doi={10.1613/jair.2502}
}

@inproceedings{carlino2013auction,
  title={Auction-Based Autonomous Intersection Management},
  author={Carlino, Dustin and Boyles, Stephen D. and Stone, Peter},
  booktitle={Proceedings of the 16th International IEEE Conference on Intelligent Transportation Systems},
  pages={529--534},
  year={2013},
  doi={10.1109/ITSC.2013.6728285}
}

@article{wang2021game,
  title={Game-Theory-Inspired Hierarchical Distributed Control Strategy for Cooperative Intersection Considering Priority Negotiation},
  author={Wang, Kaizheng and Wang, Yafei and Du, Haiping and Nam, Kanghyun},
  journal={IEEE Transactions on Vehicular Technology},
  volume={70},
  number={7},
  pages={6438--6449},
  year={2021},
  doi={10.1109/TVT.2021.3086563}
}

@article{fan2026joint,
  title={Joint Optimization of {CAV} Trajectories and Signal Phase Control in Mixed Traffic at Signalized Intersections},
  author={Fan, Ruochuan and Lu, Jian and Liu, Xian},
  journal={Journal of Transportation Engineering, Part A: Systems},
  volume={152},
  number={10},
  year={2026},
  doi={10.1061/JTEPBS.TEENG-9621}
}

@article{rios2017survey,
  title={A Survey on the Coordination of Connected and Automated Vehicles at Intersections and Merging at Highway On-Ramps},
  author={Rios-Torres, Jackeline and Malikopoulos, Andreas A.},
  journal={IEEE Transactions on Intelligent Transportation Systems},
  volume={18},
  number={5},
  pages={1066--1077},
  year={2017},
  doi={10.1109/TITS.2016.2600504}
}

@article{ntousakis2016optimal,
  title={Optimal Vehicle Trajectory Planning in the Context of Cooperative Merging on Highways},
  author={Ntousakis, Ioannis A. and Nikolos, Ioannis K. and Papageorgiou, Markos},
  journal={Transportation Research Part C: Emerging Technologies},
  volume={71},
  pages={464--488},
  year={2016},
  doi={10.1016/j.trc.2016.08.007}
}

@article{karimi2020cooperative,
  title={Cooperative Merging Control via Model Predictive Control in Mixed Traffic Environment},
  author={Karimi, Hamidreza and Liu, Chunhui and Bichiou, Youssef and Azimi, Reza},
  journal={Transportation Research Part C: Emerging Technologies},
  volume={116},
  pages={102663},
  year={2020},
  doi={10.1016/j.trc.2020.102663}
}

@article{monderer1996potential,
  title={Potential Games},
  author={Monderer, Dov and Shapley, Lloyd S.},
  journal={Games and Economic Behavior},
  volume={14},
  number={1},
  pages={124--143},
  year={1996},
  doi={10.1006/game.1996.0044}
}

@article{kiran2022deep,
  title={Deep Reinforcement Learning for Autonomous Driving: A Survey},
  author={Kiran, B. Ravi and Sobh, Ibrahim and Talpaert, Victor and Mannion, Patrick and Al Sallab, Ahmad A. and Yogamani, Senthil and Perez, Patrick},
  journal={IEEE Transactions on Intelligent Transportation Systems},
  volume={23},
  number={6},
  pages={4909--4926},
  year={2022},
  doi={10.1109/TITS.2021.3054625}
}

@article{wu2022flow,
  title={Flow: A Modular Learning Framework for Autonomy in Traffic},
  author={Wu, Cathy and Kreidieh, Abdul Rahman and Parvate, Kanaad and Vinitsky, Eugene and Bayen, Alexandre M.},
  journal={IEEE Transactions on Robotics},
  volume={38},
  number={2},
  pages={1270--1286},
  year={2022},
  doi={10.1109/TRO.2021.3087314}
}

@inproceedings{vinitsky2018benchmarks,
  title={Benchmarks for Reinforcement Learning in Mixed-Autonomy Traffic},
  author={Vinitsky, Eugene and Kreidieh, Abdul Rahman and Le Flem, Luc and Kheterpal, Nishant and Jang, Kathy and Wu, Fangyu and Wu, Cathy and Liaw, Richard and Liang, Eric and Bayen, Alexandre M.},
  booktitle={Proceedings of The 2nd Conference on Robot Learning},
  pages={399--409},
  year={2018},
  series={Proceedings of Machine Learning Research},
  volume={87}
}

@inproceedings{yu2022surprising,
  title={The Surprising Effectiveness of PPO in Cooperative Multi-Agent Games},
  author={Yu, Chao and Velu, Akash and Vinitsky, Eugene and Gao, Yu and Wang, Yu and Bayen, Alexandre and Wu, Yi},
  booktitle={Advances in Neural Information Processing Systems},
  volume={35},
  pages={24611--24624},
  year={2022}
}

@inproceedings{johannink2019residual,
  title={Residual Reinforcement Learning for Robot Control},
  author={Johannink, Tobias and Bahl, Shikhar and Nair, Ashvin and Luo, Jianlan and Kumar, Avinash and Loskyll, Matthias and Ojea, Juan Aparicio and Solowjow, Eugen and Levine, Sergey},
  booktitle={2019 International Conference on Robotics and Automation},
  pages={6023--6029},
  year={2019},
  doi={10.1109/ICRA.2019.8794127}
}

@inproceedings{alshiekh2018safe,
  title={Safe Reinforcement Learning via Shielding},
  author={Alshiekh, Mohammed and Bloem, Roderick and Ehlers, R{\"u}diger and K{\"o}nighofer, Bettina and Niekum, Scott and Topcu, Ufuk},
  booktitle={Proceedings of the AAAI Conference on Artificial Intelligence},
  volume={32},
  number={1},
  year={2018},
  doi={10.1609/aaai.v32i1.11797}
}

@article{wabersich2021predictive,
  title={A Predictive Safety Filter for Learning-Based Control of Constrained Nonlinear Dynamical Systems},
  author={Wabersich, Kim P. and Zeilinger, Melanie N.},
  journal={Automatica},
  volume={129},
  pages={109597},
  year={2021},
  doi={10.1016/j.automatica.2021.109597}
}

@article{bautista2022autonomous,
  title={Autonomous Navigation at Unsignalized Intersections: A Coupled Reinforcement Learning and Model Predictive Control Approach},
  author={Bautista-Montesano, Rolando and Galluzzi, Renato and Ruan, Kangrui and Fu, Yongjie and Di, Xuan},
  journal={Transportation Research Part C: Emerging Technologies},
  volume={139},
  pages={103662},
  year={2022},
  doi={10.1016/j.trc.2022.103662}
}

@article{wang2024multiagent,
  title={A Multi-Agent Reinforcement Learning-Based Longitudinal and Lateral Control of {CAVs} to Improve Traffic Efficiency in a Mandatory Lane Change Scenario},
  author={Wang, Shupei and Wang, Ziyang and Jiang, Rui and Zhu, Feng and Yan, Ruidong and Shang, Ying},
  journal={Transportation Research Part C: Emerging Technologies},
  volume={158},
  pages={104445},
  year={2024},
  doi={10.1016/j.trc.2023.104445}
}

@article{schwarting2018planning,
  title={Planning and Decision-Making for Autonomous Vehicles},
  author={Schwarting, Wilko and Alonso-Mora, Javier and Rus, Daniela},
  journal={Annual Review of Control, Robotics, and Autonomous Systems},
  volume={1},
  pages={187--210},
  year={2018},
  doi={10.1146/annurev-control-060117-105157}
}

@article{guanetti2018control,
  title={Control of Connected and Automated Vehicles: State of the Art and Future Challenges},
  author={Guanetti, Jacopo and Kim, Yeojun and Borrelli, Francesco},
  journal={Annual Reviews in Control},
  volume={45},
  pages={18--40},
  year={2018},
  doi={10.1016/j.arcontrol.2018.04.011}
}

@article{talebpour2016influence,
  title={Influence of Connected and Autonomous Vehicles on Traffic Flow Stability and Throughput},
  author={Talebpour, Alireza and Mahmassani, Hani S.},
  journal={Transportation Research Part C: Emerging Technologies},
  volume={71},
  pages={143--163},
  year={2016},
  doi={10.1016/j.trc.2016.07.007}
}

@article{malikopoulos2018decentralized,
  title={A Decentralized Energy-Optimal Control Framework for Connected Automated Vehicles at Signal-Free Intersections},
  author={Malikopoulos, Andreas A. and Cassandras, Christos G. and Zhang, Yue J.},
  journal={Automatica},
  volume={93},
  pages={244--256},
  year={2018},
  doi={10.1016/j.automatica.2018.03.056}
}

@article{feng2018spatiotemporal,
  title={Spatiotemporal Intersection Control in a Connected and Automated Vehicle Environment},
  author={Feng, Yiheng and Yu, Chunhui and Liu, Henry X.},
  journal={Transportation Research Part C: Emerging Technologies},
  volume={89},
  pages={364--383},
  year={2018},
  doi={10.1016/j.trc.2018.02.001}
}

@article{guo2019joint,
  title={Joint Optimization of Vehicle Trajectories and Intersection Controllers With Connected Automated Vehicles: Combined Dynamic Programming and Shooting Heuristic Approach},
  author={Guo, Yi and Ma, Jiaqi and Xiong, Chenfeng and Li, Xiaopeng and Zhou, Fang and Hao, Wei},
  journal={Transportation Research Part C: Emerging Technologies},
  volume={98},
  pages={54--72},
  year={2019},
  doi={10.1016/j.trc.2018.11.010}
}

@article{yu2019corridor,
  title={Corridor Level Cooperative Trajectory Optimization With Connected and Automated Vehicles},
  author={Yu, Chunhui and Feng, Yiheng and Liu, Henry X. and Ma, Wanjing and Yang, Xiaoguang},
  journal={Transportation Research Part C: Emerging Technologies},
  volume={105},
  pages={405--421},
  year={2019},
  doi={10.1016/j.trc.2019.06.002}
}

@article{mirheli2018signal,
  title={Development of a Signal-Head-Free Intersection Control Logic in a Fully Connected and Autonomous Vehicle Environment},
  author={Mirheli, Amir and Hajibabai, Leila and Hajbabaie, Ali},
  journal={Transportation Research Part C: Emerging Technologies},
  volume={92},
  pages={412--425},
  year={2018},
  doi={10.1016/j.trc.2018.04.026}
}

@article{mirheli2019consensus,
  title={A Consensus-Based Distributed Trajectory Control in a Signal-Free Intersection},
  author={Mirheli, Amir and Tajalli, Mehrdad and Hajibabai, Leila and Hajbabaie, Ali},
  journal={Transportation Research Part C: Emerging Technologies},
  volume={100},
  pages={161--176},
  year={2019},
  doi={10.1016/j.trc.2019.01.004}
}

@article{li2019temporal,
  title={Temporal-Spatial Dimension Extension-Based Intersection Control Formulation for Connected and Autonomous Vehicle Systems},
  author={Li, Zhenning and Wu, Qiong and Yu, Hao and Chen, Cong and Zhang, Guohui and Tian, Zong Z. and Prevedouros, Panos D.},
  journal={Transportation Research Part C: Emerging Technologies},
  volume={104},
  pages={234--248},
  year={2019},
  doi={10.1016/j.trc.2019.05.003}
}

@article{wang2020organization,
  title={Cooperative Autonomous Traffic Organization Method for Connected Automated Vehicles in Multi-Intersection Road Networks},
  author={Wang, Yunpeng and Cai, Pinlong and Lu, Guangquan},
  journal={Transportation Research Part C: Emerging Technologies},
  volume={111},
  pages={458--476},
  year={2020},
  doi={10.1016/j.trc.2019.12.018}
}

@article{zhang2019decentralized,
  title={Decentralized Optimal Control of Connected Automated Vehicles at Signal-Free Intersections Including Comfort-Constrained Turns and Safety Guarantees},
  author={Zhang, Yue and Cassandras, Christos G.},
  journal={Automatica},
  volume={109},
  pages={108563},
  year={2019},
  doi={10.1016/j.automatica.2019.108563}
}

@article{katriniok2022distributed,
  title={Fully Distributed Model Predictive Control of Connected Automated Vehicles in Intersections: Theory and Vehicle Experiments},
  author={Katriniok, Alexander and Rosarius, Benedikt and Mahonen, Petri},
  journal={IEEE Transactions on Intelligent Transportation Systems},
  volume={23},
  number={10},
  pages={18288--18300},
  year={2022},
  doi={10.1109/TITS.2022.3162038}
}

@inproceedings{lowe2017multiagent,
  title={Multi-Agent Actor-Critic for Mixed Cooperative-Competitive Environments},
  author={Lowe, Ryan and Wu, Yi and Tamar, Aviv and Harb, Jean and Abbeel, Pieter and Mordatch, Igor},
  booktitle={Advances in Neural Information Processing Systems},
  volume={30},
  pages={6379--6390},
  year={2017}
}

@inproceedings{foerster2018counterfactual,
  title={Counterfactual Multi-Agent Policy Gradients},
  author={Foerster, Jakob N. and Farquhar, Gregory and Afouras, Triantafyllos and Nardelli, Nantas and Whiteson, Shimon},
  booktitle={Proceedings of the AAAI Conference on Artificial Intelligence},
  volume={32},
  number={1},
  pages={2974--2982},
  year={2018},
  doi={10.1609/aaai.v32i1.11794}
}

@inproceedings{rashid2018qmix,
  title={{QMIX}: Monotonic Value Function Factorisation for Deep Multi-Agent Reinforcement Learning},
  author={Rashid, Tabish and Samvelyan, Mikayel and Schroeder, Christian and Farquhar, Gregory and Foerster, Jakob and Whiteson, Shimon},
  booktitle={Proceedings of the 35th International Conference on Machine Learning},
  series={Proceedings of Machine Learning Research},
  volume={80},
  pages={4295--4304},
  year={2018}
}

@article{gronauer2022survey,
  title={Multi-Agent Deep Reinforcement Learning: A Survey},
  author={Gronauer, Sven and Diepold, Klaus},
  journal={Artificial Intelligence Review},
  volume={55},
  pages={895--943},
  year={2022},
  doi={10.1007/s10462-021-09996-w}
}

@article{wong2023challenges,
  title={Deep Multiagent Reinforcement Learning: Challenges and Directions},
  author={Wong, Annie and Back, Thomas and Kononova, Anna V. and Plaat, Aske},
  journal={Artificial Intelligence Review},
  volume={56},
  number={6},
  pages={5023--5056},
  year={2023},
  doi={10.1007/s10462-022-10299-x}
}

@article{brunke2022safe,
  title={Safe Learning in Robotics: From Learning-Based Control to Safe Reinforcement Learning},
  author={Brunke, Lukas and Greeff, Melissa and Hall, Adam W. and Yuan, Zhaocong and Zhou, Siqi and Panerati, Jacopo and Schoellig, Angela P.},
  journal={Annual Review of Control, Robotics, and Autonomous Systems},
  volume={5},
  pages={411--444},
  year={2022},
  doi={10.1146/annurev-control-042920-020211}
}

@article{hewing2020learning,
  title={Learning-Based Model Predictive Control: Toward Safe Learning in Control},
  author={Hewing, Lukas and Wabersich, Kim P. and Menner, Marcel and Zeilinger, Melanie N.},
  journal={Annual Review of Control, Robotics, and Autonomous Systems},
  volume={3},
  pages={269--296},
  year={2020},
  doi={10.1146/annurev-control-090419-075625}
}

@article{narvekar2020curriculum,
  title={Curriculum Learning for Reinforcement Learning Domains: A Framework and Survey},
  author={Narvekar, Sanmit and Peng, Bei and Leonetti, Matteo and Sinapov, Jivko and Taylor, Matthew E. and Stone, Peter},
  journal={Journal of Machine Learning Research},
  volume={21},
  number={181},
  pages={1--50},
  year={2020}
}

@article{garcia2015safe,
  title={A Comprehensive Survey on Safe Reinforcement Learning},
  author={Garcia, Javier and Fernandez, Fernando},
  journal={Journal of Machine Learning Research},
  volume={16},
  number={42},
  pages={1437--1480},
  year={2015}
}

@article{fisac2019general,
  title={A General Safety Framework for Learning-Based Control in Uncertain Robotic Systems},
  author={Fisac, Jaime F. and Akametalu, Anayo K. and Zeilinger, Melanie N. and Kaynama, Shahab and Gillula, Jeremy and Tomlin, Claire J.},
  journal={IEEE Transactions on Automatic Control},
  volume={64},
  number={7},
  pages={2737--2752},
  year={2019},
  doi={10.1109/TAC.2018.2876389}
}

@article{zanon2021safe,
  title={Safe Reinforcement Learning Using Robust {MPC}},
  author={Zanon, Mario and Gros, Sebastien},
  journal={IEEE Transactions on Automatic Control},
  volume={66},
  number={8},
  pages={3638--3652},
  year={2021},
  doi={10.1109/TAC.2020.3024161}
}

@inproceedings{rusu2016policy,
  title={Policy Distillation},
  author={Rusu, Andrei A. and Colmenarejo, Sergio Gomez and Gulcehre, Caglar and Desjardins, Guillaume and Kirkpatrick, James and Pascanu, Razvan and Mnih, Volodymyr and Kavukcuoglu, Koray and Hadsell, Raia},
  booktitle={International Conference on Learning Representations},
  year={2016}
}

@inproceedings{bengio2009curriculum,
  title={Curriculum Learning},
  author={Bengio, Yoshua and Louradour, Jerome and Collobert, Ronan and Weston, Jason},
  booktitle={Proceedings of the 26th Annual International Conference on Machine Learning},
  pages={41--48},
  year={2009},
  doi={10.1145/1553374.1553380}
}

@article{chi2024spatiotemporal,
  title={Spatiotemporal-restricted {A*} algorithm as a support for lane-free traffic at intersections with mixed flows},
  author={Chi, Haifei and Cai, Pinlong and Fu, Daocheng and Zhai, Junda and Zeng, Yadan and Shi, Botian},
  journal={Green Energy and Intelligent Transportation},
  volume={3},
  number={2},
  pages={100159},
  year={2024},
  doi={10.1016/j.geits.2024.100159}
}

@article{chen2026personalized,
  title={Personalized longitudinal motion planning based on a combination of reinforcement learning and imitation learning},
  author={Chen, Chongpu and Chen, Xinbo and Hang, Peng},
  journal={Green Energy and Intelligent Transportation},
  volume={5},
  number={2},
  pages={100321},
  year={2026},
  doi={10.1016/j.geits.2025.100321}
}

@article{zeng2026reinforced,
  title={Reinforced Imitative Trajectory Planning for Urban Automated Driving},
  author={Zeng, Di and Zheng, Ling and Yang, Xiantong and Li, Yinong},
  journal={Automotive Innovation},
  year={2026},
  doi={10.1007/s42154-025-00430-3}
}

@article{wang2026feasible,
  title={Feasible Reinforcement Learning for Safety Ensurance with Stabilized Training},
  author={Wang, Zuoshuai and Lyu, Yao and Zhang, Congsheng and Xin, Zhe and Gao, Jiaxin and Jiang, Beiyan and Li, Shengbo Eben},
  journal={Automotive Innovation},
  volume={9},
  number={2},
  pages={340--353},
  year={2026},
  doi={10.1007/s42154-025-00380-w}
}

\end{document}